%% file: main.tex
\documentclass{article}

\usepackage{microtype}
\usepackage{graphicx}
\usepackage{subcaption}
\usepackage{multirow}
\usepackage{booktabs} 

\usepackage{amsmath,epsfig,bm, amssymb,graphicx,algorithm,algorithmic,color}

\input{mathdef}

\usepackage{hyperref}

\usepackage[accepted]{icml2021}

\icmltitlerunning{Submission and Formatting Instructions for ICML 2021}

\begin{document}

\twocolumn[
\icmltitle{Dynamic Sasvi: Strong Safe Screening for Norm-Regularized Least Squares}

\icmlsetsymbol{equal}{*}

\begin{icmlauthorlist}
\icmlauthor{Hiroaki Yamada}{kyoto}
\icmlauthor{Makoto Yamada}{kyoto,riken}
\end{icmlauthorlist}

\icmlaffiliation{kyoto}{Kyoto University}
\icmlaffiliation{riken}{RIKEN AIP}

\icmlcorrespondingauthor{Hiroaki Yamada}{hyamada@ml.ist.i.kyoto-u.ac.jp}
\icmlcorrespondingauthor{Makoto Yamada}{myamada@i.kyoto-u.ac.jp}

\icmlkeywords{Machine Learning, ICML}

\vskip 0.3in
]

\printAffiliationsAndNotice 

\begin{abstract}
A recently introduced technique for a sparse optimization problem called "safe screening" allows us to identify irrelevant variables in the early stage of optimization.
In this paper, we first propose a flexible framework for safe screening based on the Fenchel-Rockafellar duality and then derive a strong safe screening rule for norm-regularized least squares by the framework.
We call the proposed screening rule for norm-regularized least squares "dynamic Sasvi" because it can be interpreted as a generalization of Sasvi.
Unlike the original Sasvi, it does not require the exact solution of a more strongly regularized problem; hence, it works safely in practice.
We show that our screening rule can eliminate more features and increase the speed of the solver in comparison with other screening rules both theoretically and experimentally.
\end{abstract}

\section{Introduction}

Sparse models such as Lasso \citep{tibshirani1996regression} and group Lasso \citep{yuan2006group} have been widely studied in the areas of statistics and machine learning, and are used for various applications such as compressed sensing \citep{donoho2006compressed} and biomarker discovery \citep{climente2019block}, to name a few. Although sparse models can be formulated as a simple convex optimization problem, the computational cost can be large if the numbers of samples and dimensions are extremely large.

To tackle this problem, a technique called safe screening has been introduced \citep{ghaoui2010safe} for Lasso problems. Specifically, 
it eliminates variables that are guaranteed to be zero in the Lasso solution before solving the original Lasso optimization problem.
Many safe screening methods have been proposed for various problems \citep{ghaoui2010safe,ogawa2013safe, wang2015lasso,liu2014safe,Xiang2017screening}.
These are called sequential screening rules because they require the solution to a more strongly regularized problem.
A recent technique used to eliminate variables through an estimated solution in an iterative solver, called dynamic screening, has been proposed \citep{bonnefoy2015dynamic}.
In particular, Gap Safe \citep{fercoq2015mind, ndiaye2015gap}, a dynamic screening framework is widely used owing to its generality and efficiency \citep{ndiaye2017gap,shibagaki2016simultaneous,bao2020fast,raj2016screening,ndiaye2020screening}. More specifically, Gap Safe efficiently screens variables by using the dual form of the original problems, where the screening is characterized by properly designing the dual safe region. For Lasso, two simple region-based approaches exist: Gap Safe Sphere and Gap Safe Dome \citep{fercoq2015mind}.

In this paper, we propose a dynamic safe screening algorithm that is stronger than either Gap Safe Sphere or Gap Safe Dome for the \emph{Lasso-Like} problem, which includes norm-regularized least squares. To this end, we first propose a general screening framework based on the Fenchel-Rockafellar duality and then derive \emph{Dynamic Sasvi}, a strong safe screening rule for \emph{Lasso-like} problems.
Our framework can be regarded as a generalization of the Gap Safe framework, and thus we can derive Gap Safe Sphere and Gap Safe Dome simply using our results. Moreover, thanks to this generalization, we can use a strong problem adaptive inequality.
Interestingly, the derived screening rule for \emph{Lasso-like} problems can be seen as a dynamic variant of the safe screening with variational inequalities (Sasvi) \citep{liu2014safe}, a sequential screening rule for Lasso. Therefore, we call this dynamic Sasvi. Unlike the original Sasvi, dynamic Sasvi does not require an exact solution to the problem with another hyper-parameter and hence operates safely in practice. Moreover, we propose the use of dynamic enhanced dual polytope projections (EDPP) \citep{wang2015lasso}, which are a relaxation of dynamic Sasvi by introducing a minimum radius sphere. We show both theoretically and experimentally that the screening power and computational costs of Dynamic Sasvi and Dynamic EDPP compare favorably with those of other state-of-the-art Gap Safe methods.

\textbf{Contribution:} The contributions of our paper are summarized as follows.
\begin{itemize}
    \item We propose a flexible screening framework based on Fenchel-Rockafellar duality, which is a generalization of the Gap Safe framework \citep{ndiaye2017gap}.
    \item We propose two novel dynamic screening rules for norm-regularized least squares, which are a dynamic variant of Sasvi \citep{liu2014safe} and a dynamic variant of EDPP.
    \item We show that Dynamic Sasvi eliminates more features and increases the speed of the solver in comparison to Gap Safe \citep{fercoq2015mind,ndiaye2017gap} both theoretically and experimentally.
\end{itemize}

\section{Preliminary}
In this section, we first formulate the problem and introduce the key techniques used in this study.

\subsection{Notation}
Given $h:\mathbb{R}^m \to [-\infty,\infty]$, the domain of $h$ is defined by 
\[
\mathrm{dom}(h) := \{ \boldz\in\mathbb{R}^m \mid |h(\boldz)|<\infty \}
\]
and $h^\star:\mathbb{R}^m \to [-\infty,\infty]$, the Fenchel conjugate of $h$, is defined by
\[
h^\star(\boldv) := \sup_{\boldz\in\mathbb{R}^d} \boldv^\top\boldz - h(\boldz).
\]
If $h$ is proper, the Fenchel-Young inequality
\begin{equation}
h(\boldz) + h^\star(\boldv) \ge \boldv^\top\boldz \label{eq:fenchel_young}
\end{equation}
can be proven directly from the definition of the Fenchel conjugate.
The subdifferential of a proper function $h:\mathbb{R}^m \to (-\infty,\infty]$ at $\boldz$ is given as
\begin{align*}
& \partial h(\boldz) \\
:= & \{ \boldv\in\mathbb{R}^m \mid \forall\boldw\in\mathbb{R}^m \  \boldv^\top(\boldw-\boldz) + h(\boldz) \le h(\boldw) \}.
\end{align*}
The next proposition is important for driving Safe-screening algorithms.
\begin{prop}
\label{prop:subdiff_and_conjugate}
Assume that $h:\mathbb{R}^m\to(-\infty,\infty]$ is a proper lower semicontinuous convex function and $\boldz,\boldv\in\mathbb{R}^m$. We then have
\begin{align*}
\boldv\in\partial h(\boldz) & \iff h(\boldz) + h^\star(\boldv) = \boldv^\top\boldz \\
& \iff \boldz\in\partial h^\star(\boldv).
\end{align*}
\end{prop}
See \citep{bauschke2011convex} Section 16 for the proof.

For convex set $C\subset\mathbb{R}^m$, the relative interior of $C$ is defined by
\begin{align*}
& \mathrm{relint}(C) \\
:= & \{ v \in C \mid \forall w \in C\ \exists \epsilon > 0\ \mathrm{s.t.}\ v + \epsilon (v-w) \in C \}.
\end{align*}

\subsection{Problem Formulation}
In this study, we consider an optimization problem, formulated as
\begin{equation}
\underset{\boldbeta\in\mathbb{R}^d}{\mathrm{minimize}}~~ f(\boldX\boldbeta) + g(\boldbeta), \label{eq:primal_problem}
\end{equation}
where $\boldbeta\in\mathbb{R}^d$ is the optimization variable, $\boldX\in\mathbb{R}^{n\times d}$ is a constant matrix, and $f:\mathbb{R}^n\to(-\infty,\infty]$ and $g:\mathbb{R}^d\to(-\infty,\infty]$ are proper lower semicontinuous convex functions. We assume 
\[
\exists \boldbeta \in \mathrm{relint}(\mathrm{dom}(g)) \ \mathrm{s.t.} \  \boldX\boldbeta \in \mathrm{relint}(\mathrm{dom}(f))
\]
and the existence of the optimal point, i.e., 
\[
\exists \hat{\boldbeta} \in \mathrm{dom}(P) \ \mathrm{s.t.} \  P(\hat{\boldbeta}) = \inf_{\boldbeta\in\mathbb{R}^d} P(\boldbeta),
\]
where $P:\mathbb{R}^d\to\mathbb{R}$ is defined as $P(\boldbeta) = f(\boldX\boldbeta) + g(\boldbeta)$. Note that we have not assumed the uniqueness of the solution. Moreover, we focus on the cases where $g$ induces sparsity. Although all theorems in this paper hold, we cannot eliminate any variables without sparsity.

This class of optimization problem is popular, 
the most popular example of which is Lasso \citep{tibshirani1996regression}:
\[
\underset{\boldbeta\in\mathbb{R}^d}{\mathrm{minimize}}~~ \frac{1}{2}{\|\boldy-\boldX\boldbeta\|}_2^2 + \lambda{\|\boldbeta\|}_1.
\]
Many extensions of Lasso, including Group-Lasso \citep{yuan2006group}, Elastic-Net \citep{zou2005regularization}, and sparse logistic regression \citep{meier2008group} are in this class. Note that non-convex extensions such as SCAD \citep{fan2001variable}, Bridge \citep{frank1993statistical}, and MCP \citep{zhang2010nearly} do not satisfy this assumption.

Another example of the problem in Eq.~\eqref{eq:primal_problem} is the dual problem of a support vector machine (SVM) \citep{cortes1995support}. The dual problem of SVM can be formulated as follows: 
\[
\underset{\boldbeta\in\mathbb{R}^d : \boldzero\le\boldbeta\le\boldone}{\mathrm{minimize}}~~ \frac{1}{2}{\|\boldX\boldbeta\|}_2^2 - \boldone^\top\boldbeta.
\]
The dual problem of a support vector regression (SVR) \citep{smola2004tutorial} is also a target problem. Note that we cannot eliminate any variables of the primal problem of the normal SVM and SVR owing to a lack of sparsity. However, screening methods are available for the primal problem of the feature sparse variants of SVM and SVR \citep{ghaoui2010safe,shibagaki2016simultaneous}.

\subsection{Dual Problem}
To derive a safe screening rule for the optimization problem, Eq.~\eqref{eq:primal_problem}, the Fenchel-Rockafellar dual formulation, plays an important role.

\begin{theo} (Fenchel-Rockafellar Duality)
\label{theo:fenchel_rockafellar}
If all assumptions for the optimization problem \eqref{eq:primal_problem} are satisfied, we have the following:
\begin{equation}
\min_{\boldbeta\in\mathbb{R}^d} f(\boldX\boldbeta) + g(\boldbeta)
= \max_{\boldtheta\in\mathbb{R}^n} - f^\star(-\boldtheta) - g^\star(\boldX^\top\boldtheta). \label{eq:dual_problem}
\end{equation}
\end{theo}

The proof of Theorem \ref{theo:fenchel_rockafellar} is given in the Appendix. Let us denote $- f^\star(-\boldtheta) - g^\star(\boldX^\top\boldtheta)$ by $D(\boldtheta)$. For primal/dual solutions, we know many conditions that are equivalent to the optimality. Herein, we provide a list of such conditions for convenience.

\begin{prop} (Optimal Condition)
\label{prop:optimal_condition}
If all assumptions for the optimization problem \eqref{eq:primal_problem} are satisfied, the following are equivalent: 
\renewcommand{\labelenumi}{(\alph{enumi})}
\begin{enumerate}
    \item $\hat{\boldbeta} \in \argmin_{\boldbeta\in\mathbb{R}^d} P(\boldbeta) \land \hat{\boldtheta} \in \argmax_{\boldtheta\in\mathbb{R}^n} D(\boldtheta)$
    \item $P(\hat{\boldbeta}) = D(\hat{\boldtheta})$
    \item $f(\boldX\hat{\boldbeta}) + f^\star(-\hat{\boldtheta}) = -\hat{\boldtheta}^\top\boldX\hat{\boldbeta} = - g(\hat{\boldbeta}) - g^\star(\boldX^\top\hat{\boldtheta})$
    \item $-\hat{\boldtheta} \in \partial f(\boldX\hat{\boldbeta}) \land \boldX^\top\hat{\boldtheta} \in \partial g(\hat{\boldbeta})$
    \item $\boldX\hat{\boldbeta} \in \partial f^\star(-\hat{\boldtheta}) \land \hat{\boldbeta} \in \partial g^\star(\boldX^\top\hat{\boldtheta})$
\end{enumerate}
\end{prop}
\proof 
(a) $\iff$ (b) is directly derived from the strong duality.
(b) $\iff$ (c) is derived from the Fenchel-Young inequality \eqref{eq:fenchel_young}.
(c) $\iff$ (d) $\iff$ (e) are derived from Proposition \ref{prop:subdiff_and_conjugate}.
\proofend

\subsection{Relationship of Dual Safe Region and Screening}

In this section, we show that we can eliminate some features by constructing a simple region that contains $\hat{\boldtheta}$.

\begin{theo}
\label{theo:screening_and_dual_safe}
Assume that all assumptions for the optimization problem \eqref{eq:primal_problem} are satisfied. Let $\hat{\boldbeta}$ be the primal optimal point. Assume that the dual optimal point $\hat{\boldtheta}$ is within the region $\calR$. Then, 
\[
\hat{\boldbeta} \in \bigcup_{\boldtheta\in\calR}\partial g^\star(\boldX^\top\boldtheta).
\]
\end{theo}
\proof
According to Proposition \ref{prop:optimal_condition}, $\hat{\boldbeta} \in \partial g^\star(\boldX^\top\hat{\boldtheta}) \subset \bigcup_{\boldtheta\in\calR}\partial g^\star(\boldX^\top\boldtheta)$.
\proofend

Theorem \ref{theo:screening_and_dual_safe} provides a general method for feature screening. A simple example is the following corollary.

\begin{coro}
Consider an optimization problem, i.e., Eq.~\eqref{eq:primal_problem} with $g(\boldbeta)={\|\boldbeta\|}_1$. Assume that $\hat{\boldtheta}\in\calR$. We then have
\[
\max_{\boldtheta\in\calR}|\boldx_i^\top\boldtheta| < 1 \implies \hat{\boldbeta}_i=0.
\]
\end{coro}
\proof
By definition of $g$, we have $\partial g^\star(\boldX^\top\boldtheta) \subset \{ \boldbeta \mid \boldbeta_i=0 \} \iff |\boldx_i^\top\boldtheta| < 1$. When $\max_{\boldtheta\in\calR}|\boldx_i^\top\boldtheta| < 1$, we have $\hat{\boldbeta} \in \bigcup_{\boldtheta\in\calR}\partial g^\star(\boldX^\top\boldtheta) \subset \{ \boldbeta \mid \boldbeta_i=0 \}$ by Theorem \ref{theo:screening_and_dual_safe}.
\proofend

Note that the computational cost of $\bigcup_{\boldtheta\in\calR}\partial g^\star(\boldX^\top\boldtheta)$ depends on the simplicity of $g$ and $\calR$.

The key challenge of screening is to determine the simple narrow region $\calR$. Many regions have been proposed for various problems. In the next section, we provide a general framework for constructing a safe region.

\section{General Framework for Constructing Safe Region}

Herein, we propose a general framework for constructing a dual region that has the solution to the optimization problem in Eq.~\eqref{eq:dual_problem}.
Our framework consists of a general lower bound and a problem adaptive upper-bound of the optimal value.
Hence, we can derive a narrower region than the framework with a general upper bound under certain situations.
The general lower-bound is given in the next Theorem.

\begin{theo}
\label{theo:opt_lower_bound}
Consider the optimization problem in Eq.~\eqref{eq:dual_problem} and assume that $f^\star$ is $L$-strongly convex ($L\ge0$). Let $\hat{\boldtheta}$ be the solution to \eqref{eq:dual_problem}. Then, for $\forall \tilde{\boldtheta} \in \mathbb{R}^n$, we have
\begin{equation}
l(\hat{\boldtheta};\tilde{\boldtheta}) \le D(\hat{\boldtheta}), \label{eq:opt_lower_bound}
\end{equation}
where
\[
l(\boldtheta;\tilde{\boldtheta}) = \frac{L}{2}{\|\boldtheta-\tilde{\boldtheta}\|}_2^2 + D(\tilde{\boldtheta}).
\]
\end{theo}
\proof
According to Proposition \ref{prop:optimal_condition}, $\boldX\hat{\boldbeta} \in \partial f^\star(-\hat{\boldtheta})$ and $\hat{\boldbeta} \in \partial g^\star(\boldX^\top\hat{\boldtheta})$ hold. Because $f^\star$ is $L$-strongly convex and $g^\star$ is convex, for $\forall \tilde{\boldtheta} \in \mathbb{R}^n$, we have
\begin{align*}
f^\star(-\hat{\boldtheta}) + {(\boldX\hat{\boldbeta})}^\top(-\tilde{\boldtheta}+\hat{\boldtheta}) + \frac{L}{2}{\|\tilde{\boldtheta}-\hat{\boldtheta}\|}_2^2 & \le f^\star(-\tilde{\boldtheta}), \\
g^\star(\boldX^\top\hat{\boldtheta}) + \hat{\boldbeta}^\top(\boldX^\top\tilde{\boldtheta}-\boldX^\top\hat{\boldtheta}) & \le g^\star(\boldX^\top\tilde{\boldtheta}).
\end{align*}
Adding these two inequalities, we have the inequality \eqref{eq:opt_lower_bound}.
\proofend

This means that $\hat{\boldtheta}$ is within the region of $\{ \boldtheta \mid l(\boldtheta;\tilde{\boldtheta}) \le D(\boldtheta) \}$.
Because this region is too complicated for screening, we use a simple upper bound of $D(\boldtheta)$ to construct a simple safe region. The next theorem can be directly derived from Theorem \ref{theo:opt_lower_bound}.

\begin{theo}
\label{theo:general_safe_region}
Consider the optimization problem in Eq.~\eqref{eq:dual_problem} and assume that $f^\star$ is $L$-strongly convex ($L\ge0$). Let $\hat{\boldtheta}$ be the solution to Eq.~\eqref{eq:dual_problem}. Assume $D(\boldtheta)$ is upper bounded by $u(\boldtheta)$, i.e., $\forall \boldtheta\in\mathbb{R}^n \  D(\boldtheta) \le u(\boldtheta)$. Then, for $\forall\tilde{\boldtheta}\in\mathbb{R}^n$, we have
\[
\hat{\boldtheta} \in \mathcal{R}(\tilde{\boldtheta}, u) = \{ \boldtheta \mid l(\boldtheta;\tilde{\boldtheta}) \le u(\boldtheta) \}.
\]
\end{theo}

The complexity of $\mathcal{R}(\tilde{\boldtheta}, u)$ depends on the complexity of $u$. For example, if $u$ is linear, then $\mathcal{R}(\tilde{\boldtheta}, u)$ is a sphere.
We can construct a narrow, simple, and safe region with a tight simple upper-bound $u$.

In fact, the Gap Safe Sphere region \citep{fercoq2015mind,ndiaye2017gap} can be derived easily from this theorem and weak duality.

\begin{coro} (Gap Safe Sphere)
Consider the optimization problem in Eq.~\eqref{eq:dual_problem} and assume that $f^\star$ is $L$-strongly convex ($L\ge0$). Let $\hat{\boldtheta}$ be the solution to Eq.~\eqref{eq:dual_problem}. For $\forall\tilde{\boldbeta}\in\mathbb{R}^d$ and $\forall\tilde{\boldtheta}\in\mathbb{R}^n$, the region of the Gap Safe Sphere is given as
\begin{equation}
\calR^{\mathrm{GS}}(\tilde{\boldbeta},\tilde{\boldtheta}) = \{ \boldtheta \mid l(\boldtheta;\tilde{\boldtheta}) \le P(\tilde{\boldbeta}) \}.
\end{equation}
Then,
\[
\hat{\boldtheta} \in \calR^{\mathrm{GS}}(\tilde{\boldbeta},\tilde{\boldtheta}).
\]
\end{coro}
\proof
Based on a weak duality, we have $\forall \boldtheta \  D(\boldtheta) \le P(\tilde{\boldbeta})$. Using this constant function as an upper bound in Theorem \ref{theo:general_safe_region}, the corollary is derived directly.
\proofend

Hence, our framework can be seen as a generalization of Gap Safe.
Owing to this generalization, we can use a stronger problem-adaptive upper-bound than a weak duality. In the next section, we derive specific regions for \emph{Lasso-Like} problem. Some regions for other problems are given in the Appendix.

\section{Safe region for Lasso-like problem}

In this section, we introduce a strong upper bound for the dual problems of Lasso and similar problems. The dome region derived from it can be seen as a generalization of Sasvi \citep{liu2014safe} and is narrower than Gap Safe Sphere and Gap Safe Dome.

\subsection{Norm-regularized least squares problem and its generalization}
Norm-regularized least squares is an optimization problem and is formulated as
\[
\underset{\boldbeta\in\mathbb{R}^d}{\rm{minimize}}~~ \frac{1}{2}{\|\boldy-\boldX\boldbeta\|}_2^2 + g(\boldbeta)
\]
where $g$ is a norm. Apparently, this is a subset of problems \ref{eq:primal_problem}. Although this formulation includes Lasso \citep{tibshirani1996regression}, (overlapping) group-Lasso \citep{yuan2006group,jacob2009overlap}, and ordered weighted L1 regression \citep{figueiredo2016ordered}, the non-negative Lasso is not included. To unify them, we define the \emph{Lasso-like} problem as follows:
\begin{equation}
\underset{\boldbeta\in\mathbb{R}^d}{\mathrm{minimize}}~~ \frac{1}{2}{\|\boldy-\boldX\boldbeta\|}_2^2 + g(\boldbeta), \label{eq:primal_lassolike}
\end{equation}
where the problem satisfies all assumptions for Eq.~\eqref{eq:primal_problem} and $g$ satisfies
\begin{equation}
\forall k \ge 0, \boldbeta \in \mathbb{R}^d \  g(k\boldbeta)=kg(\boldbeta). \label{eq:g_normlike}
\end{equation}

For the Lasso-like problem, the Fenchel conjugate function of $f$ and $g$ are given as
\begin{align}
f^\star(-\boldtheta) = & \frac{1}{2}{\|\boldtheta\|}_2^2 - \boldy^\top\boldtheta, \label{eq:dual_f_sqloss} \\
g^\star(\boldX^\top\boldtheta) = & \begin{cases}
0 & (\forall\boldbeta\ \  \boldtheta^\top\boldX\boldbeta - g(\boldbeta) \le 0)\\
\infty & (\exists\boldbeta\ \  \boldtheta^\top\boldX\boldbeta - g(\boldbeta) > 0). \label{eq:dual_g_normlike}
\end{cases}
\end{align}
Note that $\{\boldtheta \mid g^\star(\boldX^\top\boldtheta)=0\}$ is a closed convex set. Hence, the Lasso-like problem is a class of problems whose Fenchel-Rockafellar dual can be seen as a convex projection.

\subsection{Proposed Dome Region for Lasso-like problem}
Thanks to Theorem \ref{theo:opt_lower_bound}, we can construct a safe region by proposing an upper bound $u(\boldtheta)$. In this section, we propose a tight upper bound for Lasso-like problems.

The direct expression of $f^\star$ in Eq.~\eqref{eq:dual_f_sqloss} is sufficiently simple. We only need an upper bound of $-g^\star$ to construct a simple region. The upper bound is given as follows: 

\begin{lemm}
\label{lemm:upper_gdual_lassolike}
For Lasso-like problems \eqref{eq:primal_lassolike}, for $\forall\tilde{\boldbeta}\in\mathbb{R}^d$ and $\forall\boldtheta\in\mathbb{R}^n$, we have
\begin{align}
- g^\star(\boldX^\top\boldtheta) & \le \inf_{k\ge0}g(k\tilde{\boldbeta}) - \boldtheta^\top\boldX(k\tilde{\boldbeta}) \nonumber \\
& = \begin{cases}
0 & (g(\tilde{\boldbeta}) - \boldtheta^\top \boldX\tilde{\boldbeta} \ge 0) \\
-\infty & (g(\tilde{\boldbeta}) - \boldtheta^\top \boldX\tilde{\boldbeta} < 0). \\
\end{cases} \label{eq:gdual_linear_const}
\end{align}
\end{lemm}
\proof
Based on a Fenchel-Young inequality \eqref{eq:fenchel_young}, we have 
\[
- g^\star(\boldX^\top\boldtheta) \le  \inf_{k\ge0} g(k\tilde{\boldbeta}) - \boldtheta^\top\boldX(k\tilde{\boldbeta}).
\]
Under the condition of Eq.~\eqref{eq:g_normlike}, we have $g(k\tilde{\boldbeta}) = k g(\tilde{\boldbeta})$. Therefore, the optimal value of the upper bound is zero if $g(\tilde{\boldbeta}) - \boldtheta^\top\boldX(\tilde{\boldbeta}) \ge 0$ and $-\infty$ otherwise.
\proofend

The next theorem can be directly derived from Lemma \ref{lemm:upper_gdual_lassolike}.
\begin{theo}
Consider Lasso-like problems in Eq.~\eqref{eq:primal_lassolike}. Let
\begin{equation}
u^{\mathrm{DS}}(\boldtheta;\tilde{\boldbeta}) := \begin{cases}
-f^\star(-\boldtheta) & (g(\tilde{\boldbeta}) - \boldtheta^\top\boldX\tilde{\boldbeta} \ge 0) \\
- \infty & (g(\tilde{\boldbeta}) - \boldtheta^\top\boldX\tilde{\boldbeta} < 0)
\end{cases}. \label{eq:upper_dsasvi}
\end{equation}
Then, for $\forall\tilde{\boldbeta}\in\mathbb{R}^d$ and $\forall\boldtheta\in\mathbb{R}^n$,
\[
D(\boldtheta) \le u^{\mathrm{DS}}(\boldtheta;\tilde{\boldbeta}).
\]
\end{theo}

Then, Theorem \ref{theo:general_safe_region} provides a simple and safe region.

\begin{theo}
\label{theo:region_dsasvi}
Consider the Lasso-like problem in Eq.~\eqref{eq:primal_lassolike} and its Fenchel-Rockafellar dual problem in Eq.~\eqref{eq:dual_problem}. Let $\hat{\boldtheta}$ be the dual optimal point. We assume that $\tilde{\boldbeta}\in\mathbb{R}^d$ and $\tilde{\boldtheta}\in\mathrm{dom}(D)$. Then, $\hat{\boldtheta}$ is within the Dynamic Sasvi region, which is given as an intersection of a sphere and a half space:
\begin{align*}
& \calR^{\mathrm{DS}}(\tilde{\boldbeta},\tilde{\boldtheta}) \\
:= & \{ \boldtheta \mid l(\boldtheta;\tilde{\boldtheta}) \le u^{\mathrm{DS}}(\boldtheta;\tilde{\boldbeta}) \}, \\
= & \{ \boldtheta \mid {\left\|\boldtheta - \frac{1}{2}(\tilde{\boldtheta}+\boldy)\right\|}_2^2 \le \frac{1}{4}{\|(\tilde{\boldtheta}-\boldy)\|}_2^2 \\. 
& \land 0 \le g(\tilde{\boldbeta}) - \boldtheta^\top\boldX\tilde{\boldbeta} \}.
\end{align*}
\end{theo}

The proof of Theorem \ref{theo:region_dsasvi} is given in the Appendix. Because of continuity, $\calR^{\mathrm{DS}}(\tilde{\boldbeta}^{(t)},\tilde{\boldtheta}^{(t)})$ converges to $\calR^{\mathrm{DS}}(\hat{\boldbeta},\hat{\boldtheta})=\{\hat{\boldtheta}\}$ if $\lim_{t\to\infty}\tilde{\boldbeta}^{(t)}=\hat{\boldbeta}$ and $\lim_{t\to\infty}\tilde{\boldtheta}^{(t)}=\hat{\boldtheta}$ hold.

\subsection{Relation to Sasvi}
In this section, we show that safe screening with variational inequality (Sasvi) \citep{liu2014safe} is a special case of our screening rule.
First, we review Sasvi. The target task of Sasvi is to minimize $\frac{1}{2}{\left\|\boldy-\boldX\boldbeta\right\|}_2^2 + \lambda{\|\boldbeta\|}_1$ with many $\lambda$s. Divided by $\lambda^2$ and change optimization variable, we obtain the following:
\[
\underset{\boldbeta\in\mathbb{R}^d}{\mathrm{minimize}}~~ \frac{1}{2}{\left\|\frac{1}{\lambda}\boldy-\boldX\boldbeta\right\|}_2^2 + {\|\boldbeta\|}_1.
\]
Let ${\hat\boldbeta}^{(\lambda)}$ and ${\hat\boldtheta}^{(\lambda)}$ be the optimal points of the primal problem and the Fenchel-Rockafellar dual problem, respectively.
Sasvi uses ${\hat\boldtheta}^{(\lambda_0)}$ to construct a safe region for ${\hat\boldtheta}^{(\lambda)}$.
Although Sasvi was originally proposed for Lasso, it can be easily generalized for the Lasso-like problem as follows.

\begin{theo}
\label{theo:sasvi_lassolike}
Let ${\hat\boldtheta}^{(\lambda)}$ be the optimal point of the Fenchel-Rockafellar dual problem of the Lasso-like problem (that is, $g$ satisfies Eq.~\eqref{eq:g_normlike}) 
\[
\underset{\boldtheta : g^\star(\boldX^\top\boldtheta)=0}{\mathrm{maximize}}~~ - \frac{1}{2}{\|\boldtheta-\frac{1}{\lambda}\boldy\|}_2^2 + \frac{1}{2}{\|\frac{1}{\lambda}\boldy\|}_2^2.
\]
Assume we have an exact ${\hat\boldtheta}^{(\lambda_0)}$.
We then have
\begin{align*}
{\hat\boldtheta}^{(\lambda)} \in & \calR^{\mathrm{Sasvi}}(\lambda,\lambda_0) \\
:= & \{ \boldtheta \mid 0 \ge {\left(\frac{1}{\lambda}\boldy-\boldtheta\right)}^\top\left({\hat\boldtheta}^{(\lambda_0)}-\boldtheta\right) \\
& \land 0 \ge {\left(\frac{1}{\lambda_0}\boldy-{\hat\boldtheta}^{(\lambda_0)}\right)}^\top\left(\boldtheta-{\hat\boldtheta}^{(\lambda_0)}\right) \}
\end{align*}
\end{theo}
\proof
Because the duality of the Lasso-like problem can be interpreted as a projection from $\frac{1}{\lambda}\boldy$ to a closed convex set $\{\boldtheta \mid g^\star(\boldX^\top\boldtheta)=0\}$, two variational inequalities hold. See \citep{liu2014safe} for more details.
\proofend

We can then prove that $\calR^{\mathrm{Sasvi}}(\lambda_0):=\calR^{\mathrm{Sasvi}}(1,\lambda_0)$ equals $\calR^{\mathrm{DS}}({\hat\boldbeta}^{(\lambda_0)},{\hat\boldtheta}^{(\lambda_0)})$.
Note that we can set $\lambda=1$ without a loss of generality because multiplying the same scalar to $\boldy$, $\lambda$, and $\lambda_0$ does not change the problem or the region.
\begin{theo}
\label{theo:equality_sasvi}
Consider the Lasso-like problem
\[
\underset{\boldbeta\in\mathbb{R}^d}{\mathrm{minimize}}~~ \frac{1}{2}{\left\|\frac{1}{\lambda}\boldy-\boldX\boldbeta\right\|}_2^2 + g(\boldbeta).
\]
Let ${\hat\boldbeta}^{(\lambda)}$ and ${\hat\boldtheta}^{(\lambda)}$ be the primal/dual optimal points, respectively. We then have
\[
\calR^{\mathrm{Sasvi}}(\lambda_0) = \calR^{\mathrm{DS}}({\hat\boldbeta}^{(\lambda_0)},{\hat\boldtheta}^{(\lambda_0)}).
\]
where $\calR^{\mathrm{Sasvi}}(\lambda_0)$ and $\calR^{\mathrm{DS}}({\hat\boldbeta}^{(\lambda_0)},{\hat\boldtheta}^{(\lambda_0)})$ are safe regions for ${\hat\boldtheta}^{(1)}$.
\end{theo}

The proof of Theorem \ref{theo:equality_sasvi} is given in the Appendix.
For this reason, we have labeled it "Dynamic Sasvi."
This generalization increases the speed of the solver significantly because the region of our method may be extremely narrow in the late stage of optimization.
As pointed out in \citep{fercoq2015mind}, some sequential safe screening rules, including Sasvi, are not safe in practice because we do not have the exact solution for $\lambda_0$.
Dynamic Sasvi overcomes this problem because its region is safe if it is not the exact solution.

\subsection{Comparison to Gap Safe Dome and Gap Safe Sphere}

\begin{figure*}[t]
  \centering
  \begin{subfigure}[t]{0.325\textwidth}
    \centering
    \includegraphics[width=0.99\textwidth]{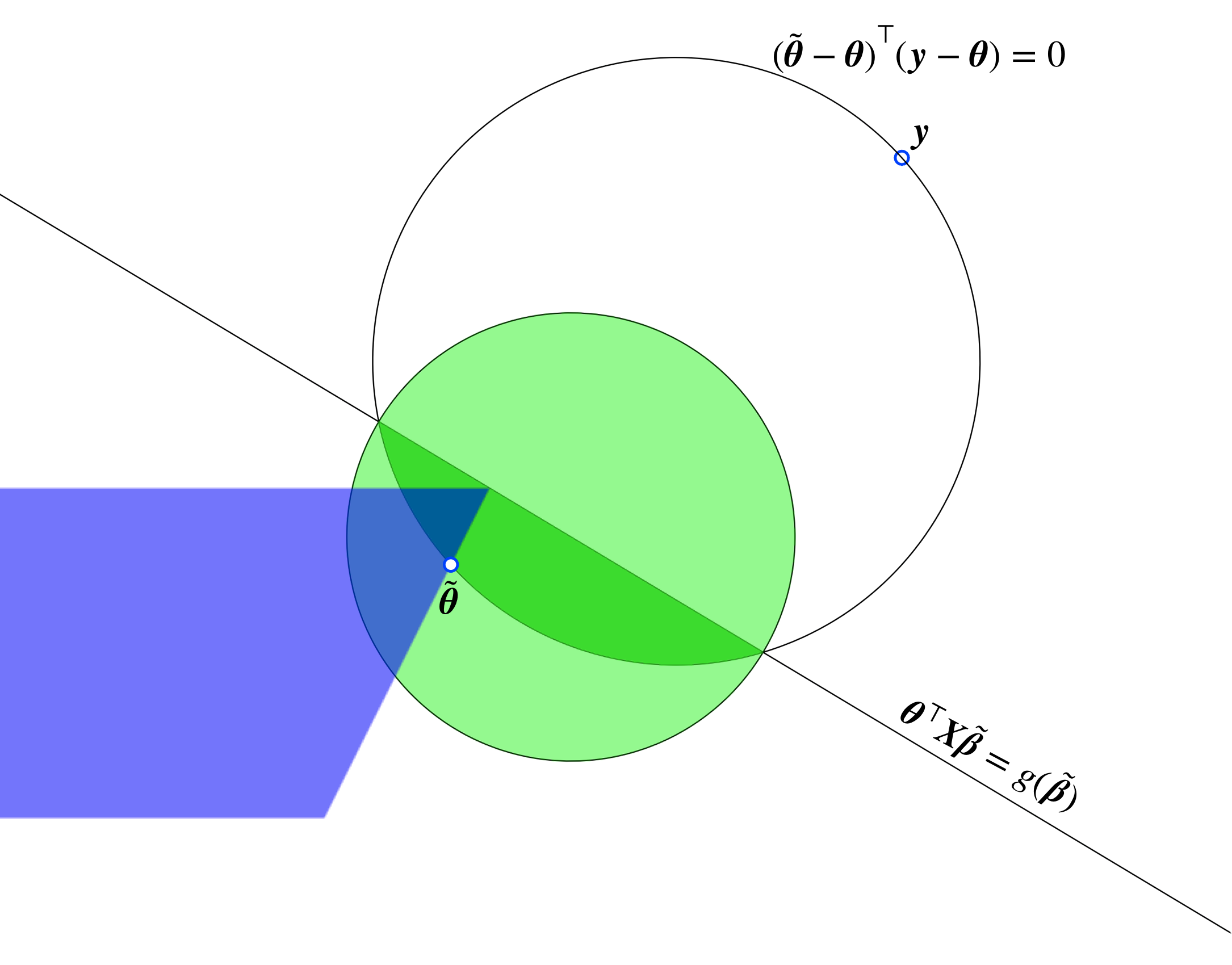}
    \caption{Regions of dynamic Sasvi (dark green) and dynamic EDPP (light green).}
    \label{fig:regions_dsde}
  \end{subfigure}
  \begin{subfigure}[t]{0.325\textwidth}
    \centering
    \includegraphics[width=0.99\textwidth]{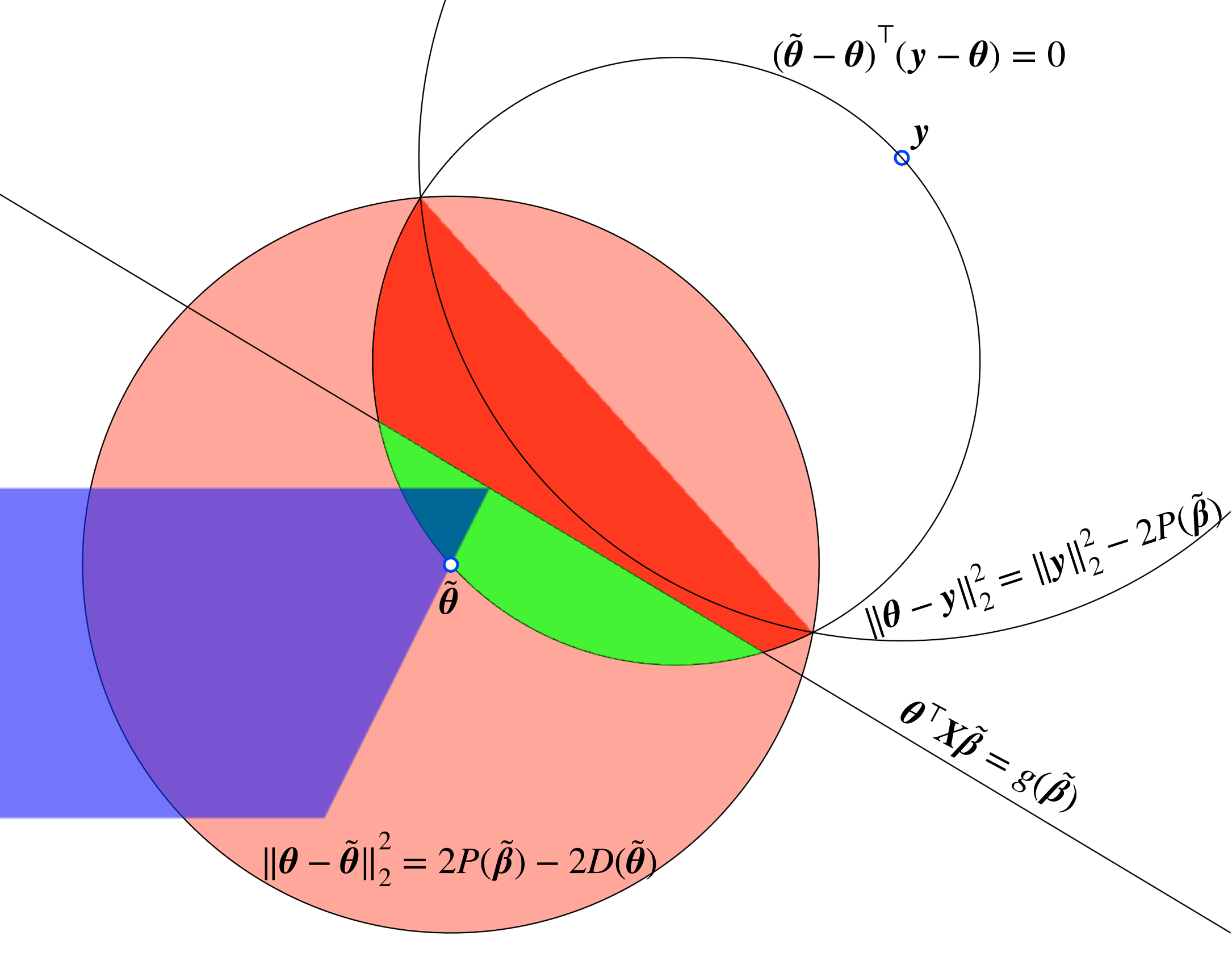}
    \caption{Regions of dynamic Sasvi (green), Gap Safe Sphere (light red) and Gap Safe Dome (dark red).}
    \label{fig:regions_dsgsgd}
  \end{subfigure}
  \begin{subfigure}[t]{0.325\textwidth}
    \centering
    \includegraphics[width=0.99\textwidth]{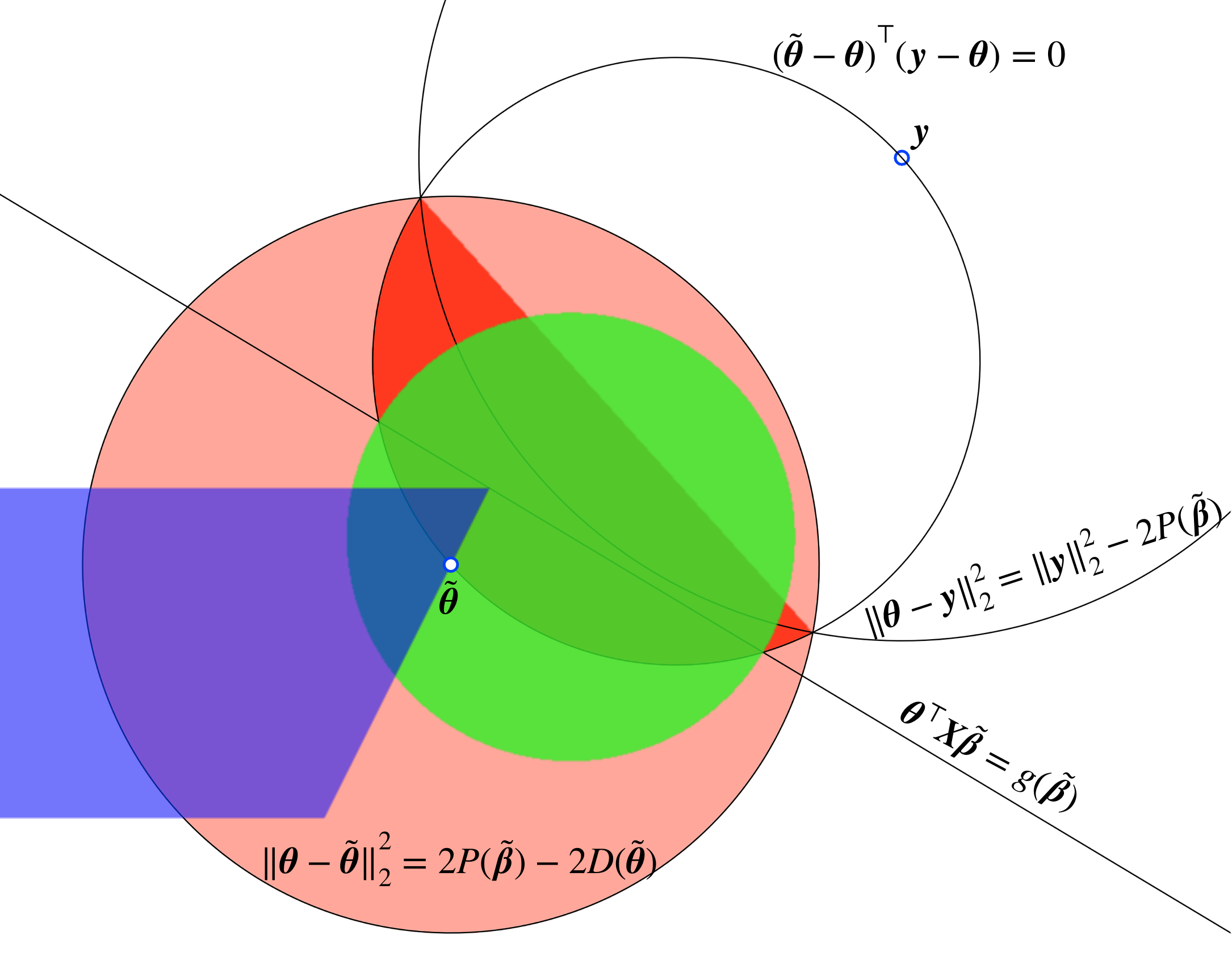}
    \caption{Regions of dynamic EDPP (green), Gap Safe Sphere (light red), Gap Safe Dome (dark red).}
    \label{fig:regions_degsgd}
  \end{subfigure}
\caption{Comparisons of various safe regions for Lasso ($\boldX=\begin{pmatrix} 2 & 0 \\ -1 & 3 \\ \end{pmatrix}$, $\boldy=\begin{pmatrix} 1.5 \\ 1 \\ \end{pmatrix}$). The blue region is the feasible region. $\tilde{\boldbeta}$ was obtained by a cycle of coordinate descent. $\tilde{\boldtheta}=\phi(\tilde{\boldbeta})$.}
  \vspace{-.2in}
 \end{figure*}

Here, we show that the proposed method is stronger than Gap Safe Dome \citep{fercoq2015mind} and Gap Safe Sphere \citep{fercoq2015mind}, \citep{ndiaye2017gap} for Lasso-like problems.
As shown in \citep{fercoq2015mind}, for Lasso, the regions of the Gap Safe Dome and Gap Safe Sphere are the relaxation of the intersection of a sphere and the contra of another sphere.
We call this unrelaxed region Gap Safe Moon.
Although Gap Safe Moon is defined only for Lasso in \citep{fercoq2015mind}, it can be naturally generalized for Lasso-like problems. Gap Safe Moon can be derived from Corollary \ref{theo:general_safe_region}.

\begin{theo} (Gap Safe Moon)
\label{theo:upperbound_gm}
Consider the Lasso-like problem in Eq.~\eqref{eq:primal_lassolike} and its dual Fenchel-Rockafellar equation, i.e., Eq. ~\eqref{eq:dual_problem}. Let $\hat{\boldtheta}$ be the dual optimal point. For $\tilde{\boldbeta}\in\mathbb{R}^d$, the Gap Safe Moon upper bound is given as
\begin{equation}
u^{\mathrm{GM}}(\boldtheta; \tilde{\boldbeta}) = \begin{cases}
-f^\star(-\boldtheta) & (-f^\star(-\boldtheta) \le P(\tilde{\boldbeta})) \\
-\infty & (-f^\star(-\boldtheta) > P(\tilde{\boldbeta}))
\end{cases}. \label{eq:upper_gm}
\end{equation}
Then, for $\forall\tilde{\boldbeta}\in\mathbb{R}^d$, $\forall\tilde{\boldtheta}\in\mathrm{dom}(D)$, and $\forall\boldtheta\in\mathbb{R}^n$, we have
\[
D(\boldtheta) \le u^{\mathrm{GM}}(\boldtheta; \tilde{\boldbeta})
\]
and hence
\begin{align*}
\hat{\boldtheta} \in & \{ \boldtheta \mid l(\boldtheta; \tilde{\boldtheta}) \le u^{\mathrm{GM}}(\boldtheta; \tilde{\boldbeta}) \} \\
= & \{ \boldtheta \mid -f^\star(-\boldtheta) \le P(\tilde{\boldbeta}) \land {(\tilde{\boldtheta}-\boldtheta)}^\top(\boldy-\boldtheta) \le 0\}.
\end{align*}
\end{theo}

The proof of Theorem \ref{theo:upperbound_gm} is given in the Appendix. We can then derive the next theorem.

\begin{theo} (Gap Safe Moon and Dynamic Sasvi)
For $\forall\tilde{\boldbeta}\in\mathbb{R}^d$ and $\forall\boldtheta\in\mathbb{R}^n$, we have
\[
u^{\mathrm{DS}}(\boldtheta; \tilde{\boldbeta}) \le u^{\mathrm{GM}}(\boldtheta; \tilde{\boldbeta}).
\]
\end{theo}
\proof
If $g(\tilde{\boldbeta}) - \boldtheta^\top\boldX\tilde{\boldbeta}$ is negative, $u^{\mathrm{DS}}(\boldtheta; \tilde{\boldbeta}) = -\infty$, and thus the inequality holds. If $0 \le g(\tilde{\boldbeta}) - \boldtheta^\top\boldX\tilde{\boldbeta}$, by adding the Fenchel-Young inequality \eqref{eq:fenchel_young}, we have $-f^\star(-\boldtheta) \le P(\tilde{\boldbeta})$ and $u^{\mathrm{DS}}(\boldtheta; \tilde{\boldbeta}) = u^{\mathrm{GM}}(\boldtheta; \tilde{\boldbeta}) = -f^\star(-\boldtheta)$.
\proofend

This theorem means that the region of dynamic Sasvi is a subset of the region of Gap Safe Moon.
Because Gap Safe Dome and Gap Safe Sphere are based on the relaxation of the Gap Safe Moon region, our screening is always stronger than them. Figure \ref{fig:regions_dsgsgd} shows the regions of Dynamic Sasvi, Gap Safe Dome and Gap Safe Sphere.

\subsection{Sphere Relaxation (Dynamic EDPP)}
In some situations, even a dome region is too complicated to calculate $\bigcup_{\boldtheta\in\calR}\partial g^\star(-\boldX^\top\boldtheta)$. We propose using a minimum radius sphere that includes the dynamic Sasvi region in such cases.
This method can be seen as a dynamic variant of the enhanced dual polytope projections (EDPP) \citep{wang2015lasso} because the EDPP is the minimum radius sphere relaxation of Sasvi.

\begin{theo}
\label{theo:region_dynamicedpp}
Consider the Lasso-like problem in Eq.~\eqref{eq:primal_lassolike} and its Fenchel-Rockafellar dual problem in Eq.~\eqref{eq:dual_problem}. We assume that $\tilde{\boldbeta}\in\mathbb{R}^d$ and $\tilde{\boldtheta}\in\mathrm{dom}(D)$. If $n\ge2$, the minimum radius sphere including $\calR^{\mathrm{DS}}(\tilde{\boldbeta},\tilde{\boldtheta})$ is 
\begin{equation}
\calR^{\mathrm{DE}}(\tilde{\boldbeta},\tilde{\boldtheta}) = \{ \boldtheta \mid {\|\boldtheta - \boldtheta_c\|}_2^2 \le r^2 \}, \label{eq:region_SLL}
\end{equation}
where 
\begin{align*}
\boldtheta_c & = \frac{1}{2}(\tilde{\boldtheta}+\boldy)-\alpha\boldX\tilde{\boldbeta} \\
r^2 & = \frac{1}{4}{\|\tilde{\boldtheta}-\boldy\|}_2^2 - \alpha^2{\|\boldX\tilde{\boldbeta}\|}_2^2 \\
\alpha & = \max\left(0, \frac{1}{{\|\boldX\tilde{\boldbeta}\|}_2^2}\left(\frac{1}{2}{(\tilde{\boldtheta}+\boldy)}^\top\boldX\tilde{\boldbeta} - g(\tilde{\boldbeta})\right)\right).
\end{align*}
\end{theo}

The proof of Theorem \ref{theo:region_dynamicedpp} is given in the Appendix. Figures \ref{fig:regions_dsde} and \ref{fig:regions_degsgd} show the dynamic EDPP region and other regions. Note that the dynamic EDPP region is not guaranteed to be within the Gap Safe Sphere region. However, its radius is always smaller than that of Gap Safe Sphere.

\section{Implementation for Lasso}

\begin{algorithm}[tb]
   \caption{Coordinate descent with Dynamic Sasvi for Lasso}
   \label{alg:implementation}
\begin{algorithmic}[1]
   \STATE {\bfseries Input: $\boldX, \boldy, \boldbeta^0, T, c, \epsilon$} 
   \STATE Initialize $\tilde{\boldbeta} \leftarrow \boldbeta^0$, $\calA \leftarrow [\![d]\!]$
   \FOR{$t \in [\![T]\!]$}
   \IF{$k \mod c = 1$}
   \STATE Compute $\tilde{\boldtheta}=\phi_\calA(\tilde{\boldbeta})$
   \IF{$P(\tilde{\boldbeta})-D(\tilde{\boldtheta}) \le \epsilon$}
   \STATE {\bf break}
   \ENDIF
   \STATE $\calR \leftarrow \calR^{\mathrm{DS}}(\tilde{\boldbeta},\tilde{\boldtheta})$
   \STATE $\calA \leftarrow \{j \in \calA : \max_{\boldtheta\in\calR}|\boldx_j^\top\boldtheta| \ge 1\}$
   \FOR{$j \in [\![d]\!]-\calA$}
   \STATE $\tilde{\boldbeta}_j \leftarrow 0$
   \ENDFOR
   \ENDIF
   \FOR{$j \in \calA$}
   \STATE $u \leftarrow \tilde{\boldbeta}_j{\|\boldx_j\|}_2^2-\boldx_j^\top(\boldX\tilde{\boldbeta}-\boldy)$
   \STATE $\tilde{\boldbeta}_j \leftarrow \frac{1}{{\|\boldx_j\|}_2^2}\text{sign}(u)\max(0,|u|-1)$
   \ENDFOR
   \ENDFOR
   \STATE {\bfseries Output:$\tilde{\boldbeta}$}
\end{algorithmic}
\end{algorithm}

In this section, we provide a specific solver based on Theorem \ref{theo:region_dsasvi}. Because the algorithm used to calculate $\bigcup_{\boldtheta\in\calR}\partial g^\star(\boldX^\top\boldtheta)$ depends on $g$, we introduce a Lasso solver as an example.
We must choose an iterative solver to combine with screening methods because they cannot estimate the solution alone. Although our methods can work with any iterative method, we use coordinate descent, which is recommended in \citep{friedman2007}.

\subsection{Choice of $\tilde{\boldtheta}$}
As shown in the previous section, $\lim_{t\to\infty}\calR^{\mathrm{DS}}(\boldbeta^t,\boldtheta^t)$ converges to $\{\hat{\boldtheta}\}$ when $\lim_{t\to\infty}\boldbeta^t=\hat{\boldbeta}$ and $\lim_{t\to\infty}\boldtheta^t=\hat{\boldtheta}$ holds.
Because the iterative solver provides such a sequence of primal points and screening does not harm its convergence, we only need a converging sequence of dual points to obtain a converging safe region.
The next theorem provides such a sequence.

\begin{theo} (Converging $\boldtheta^t$)
Consider the optimization problem Eq. \eqref{eq:primal_lassolike} with $g(\boldbeta)={\|\boldbeta\|}_1$. Let $\hat{\boldbeta}\in\mathbb{R}^d$ and $\hat{\boldtheta}\in\mathbb{R}^n$ be the primal/dual solution. Assume $\lim_{t\to\infty}\boldbeta^t=\hat{\boldbeta}$. Let us define $\phi:\mathbb{R}^d\to\mathbb{R}^n$ as
\[
\phi(\boldbeta) := \frac{1}{\max(1,{\|\boldX^\top(\boldy-\boldX\boldbeta)\|}_\infty)}(\boldy-\boldX\boldbeta).
\]
Then, $\forall \boldbeta \ \phi(\boldbeta)\in\mathrm{dom}(D)$ and $\lim_{t\to\infty}\phi(\boldbeta^t)=\hat{\boldtheta}$ hold.
\end{theo}
\proof
$\phi(\boldbeta)\in\mathrm{dom}(D)$ is directly derived from ${\|\boldX^\top\phi(\boldbeta)\|}_\infty = \min({\|\boldX^\top(\boldy-\boldX\boldbeta)\|}_\infty,1) \le 1$. Because $\phi$ is continuous and $\phi(\hat{\boldbeta})=\hat{\boldtheta}$, $\lim_{t\to\infty}\phi(\boldbeta^t)=\hat{\boldtheta}$ also holds.
\proofend

Actually, if $\calA$ is the set of features that is not yet eliminated, we can use
\[
\phi_\calA(\boldbeta) := \frac{1}{\max(1,\max_{j\in\calA}\boldx_j^\top(\boldy-\boldX\boldbeta))}(\boldX\boldbeta-\boldy)
\]
instead of $\phi(\boldbeta)$. Although $\phi_\calA(\boldbeta)\in\mathrm{dom}(D)$ is not guaranteed, $\phi_\calA(\boldbeta)$ is guaranteed to satisfy all constraints that are active in the dual solution. In other words, $\phi_\calA(\boldbeta)$ is in the domain of the dual problem of the small primal problem without eliminated features.

Now, we can optimize the problem with the proposed screening. The pseudo code is described in Algorithm \ref{alg:implementation}. Direct expression of $\max_{\boldtheta\in\calR^{\mathrm{DS}}(\tilde{\boldbeta},\tilde{\boldtheta})}|\boldx_j^\top\boldtheta|$ is given in the Appendix.

\subsection{Computational Cost of Dynamic Sasvi Screening}
In Dynamic Sasvi screening, the calculation of $\phi_\calA(\tilde{\boldbeta})$ and $\max_{\boldtheta\in\calR}|\boldx_j^\top\boldtheta|$ controls the computational cost.
If we have $\boldX\tilde{\boldbeta}$, $\boldX^\top\boldX\tilde{\boldbeta}$, and $\boldX^\top\boldy$, we can obtain $\phi_\calA(\boldbeta)$ with $O(n+d)$ calculations.
If we have $\boldX\tilde{\boldbeta}$, $\boldX^\top\boldX\tilde{\boldbeta}$, $\tilde{\boldtheta}$, $\boldX^\top\tilde{\boldtheta}$, and $\boldX^\top\boldy$, we can obtain $\max_{\boldtheta\in\calR}|\boldx_j^\top\boldtheta|$ for all $j$ with $O(n+d)$ calculations.
Because $\boldX^\top\boldy$ is constant and $\tilde{\boldtheta}=\phi_\calA(\tilde{\boldbeta})$ is a linear combination of $\boldX\tilde{\boldbeta}$ and $\boldy$, the calculations of only $\boldX\tilde{\boldbeta}$ and $\boldX^\top\boldX\tilde{\boldbeta}$ cost $O(nd)$.
Hence, the screening cost is almost the same for all methods, which require $\boldX^\top\boldX\tilde{\boldbeta}$, including Gap Safe.

\subsection{Computation of Lasso Path}
In practice, we formulate the Lasso problem as follows:
\[
\underset{\boldbeta\in\mathbb{R}^d}{\rm{minimize}}~~ \frac{1}{2}{\left\|\frac{1}{\lambda}\boldy-\boldX\boldbeta\right\|}_2^2 + {\|\boldbeta\|}_1
\]
and solve for many values of $\lambda$ to choose the best solution.
Considering the situation in which we have to estimate the solutions $\hat{\boldbeta}^{(\lambda_1)}, \hat{\boldbeta}^{(\lambda_2)}, \cdots, \hat{\boldbeta}^{(\lambda_M)}$ corresponds to $\lambda_1 > \lambda_2 > \cdots > \lambda_M$.
Many studies (e.g., \citep{fercoq2015mind}) recommend using the estimated solution for $\lambda_{m-1}$ as the initial vector in the estimation of $\hat{\boldbeta}^{(\lambda_m)}$ because $\hat{\boldbeta}^{(\lambda_{m-1})}$ and $\hat{\boldbeta}^{(\lambda_m)}$ may be close.
In our implementation, we set the initial vector as $k\tilde{\boldbeta}^{(\lambda_{m-1})}$, where $\tilde{\boldbeta}^{(\lambda_{m-1})}$ is the estimation of $\hat{\boldbeta}^{(\lambda_{m-1})}$ and
\begin{align*}
k := & \argmin_{k\ge0}\frac{1}{2}{\left\|\frac{1}{\lambda_m}\boldy-k\boldX\tilde{\boldbeta}^{(\lambda_{m-1})}\right\|}_2^2 + k{\|\tilde{\boldbeta}^{(\lambda_{m-1})}\|}_1 \\
= & \frac{1}{{\|\boldX\tilde{\boldbeta}^{(\lambda_{m-1})}\|}_2^2}({\|\tilde{\boldbeta}^{(\lambda_{m-1})}\|}_1 - \frac{1}{\lambda_m}\boldy^\top\boldX\tilde{\boldbeta}^{(\lambda_{m-1})}).
\end{align*}

\begin{figure*}[t]
  \centering
  \begin{subfigure}[t]{0.325\textwidth}
    \centering
    \includegraphics[width=0.99\textwidth]{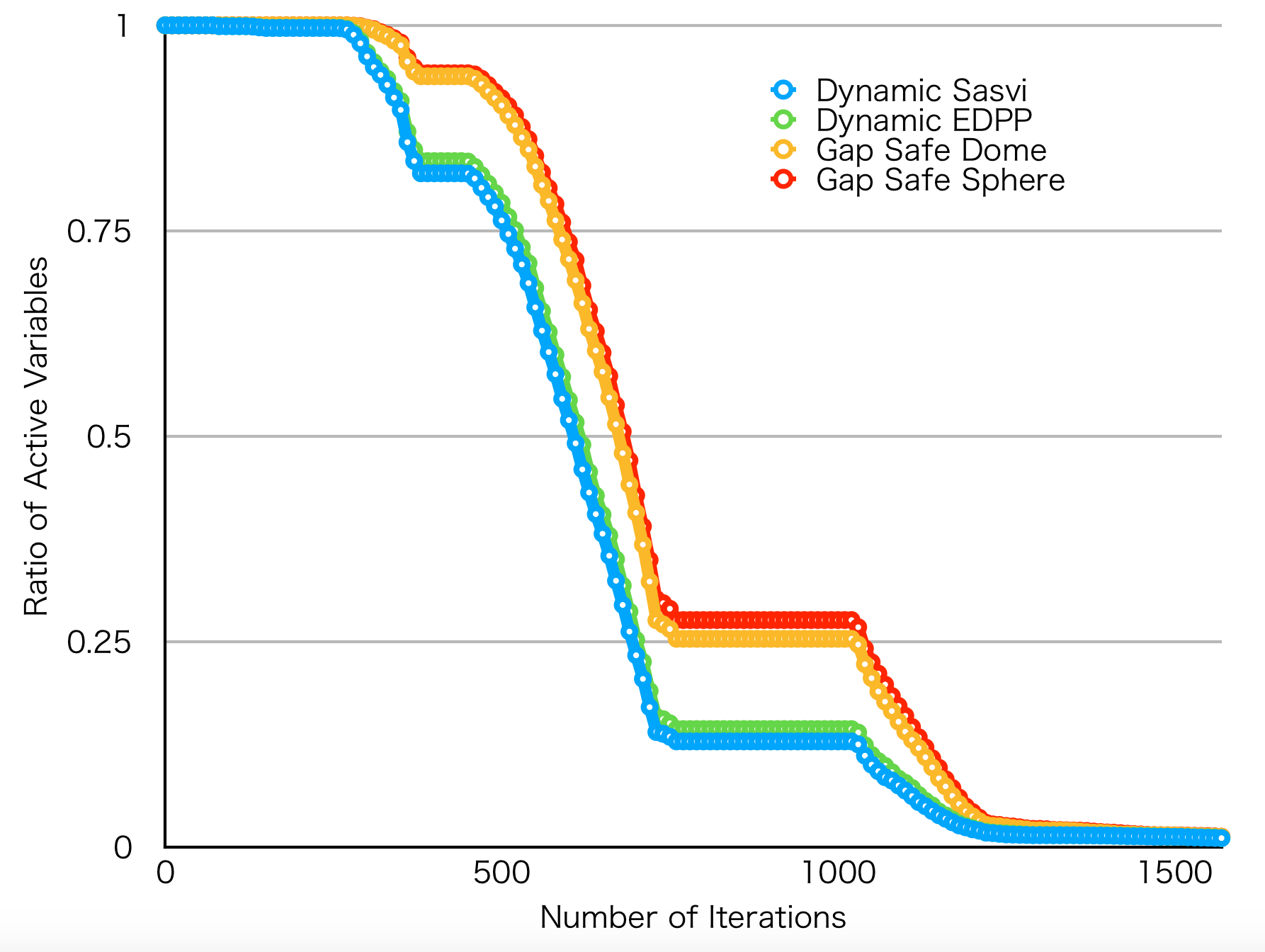}
    \caption{Feature remaining rate (Leukemia).}
    \label{fig:feature_remaininf_rate_leukemia}
  \end{subfigure}
  \begin{subfigure}[t]{0.325\textwidth}
    \centering
    \includegraphics[width=0.99\textwidth]{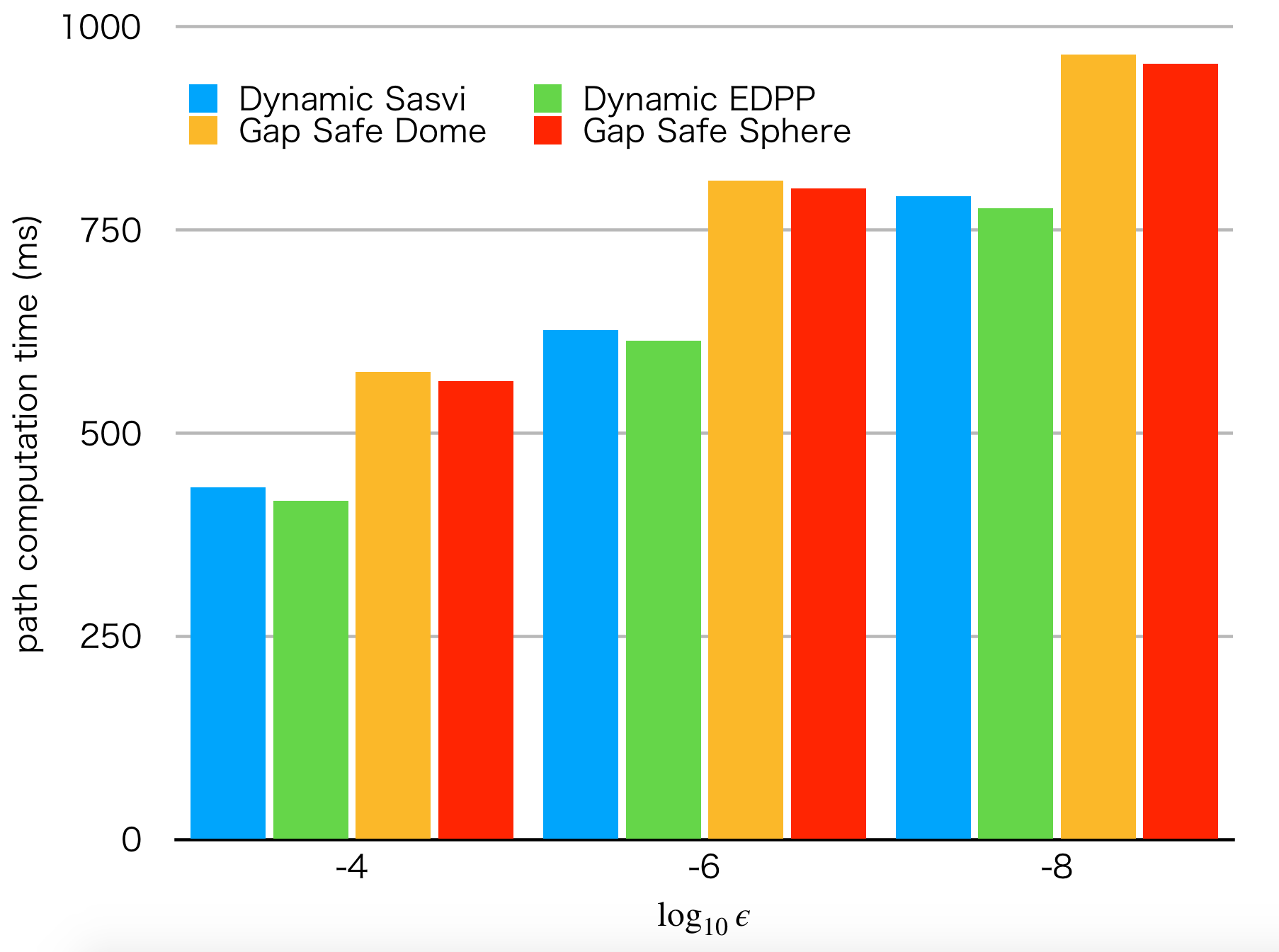}
    \caption{Computational time (Leukemia).}
    \label{fig:time_path_leukemia}
  \end{subfigure}
  \begin{subfigure}[t]{0.325\textwidth}
    \centering
    \includegraphics[width=0.99\textwidth]{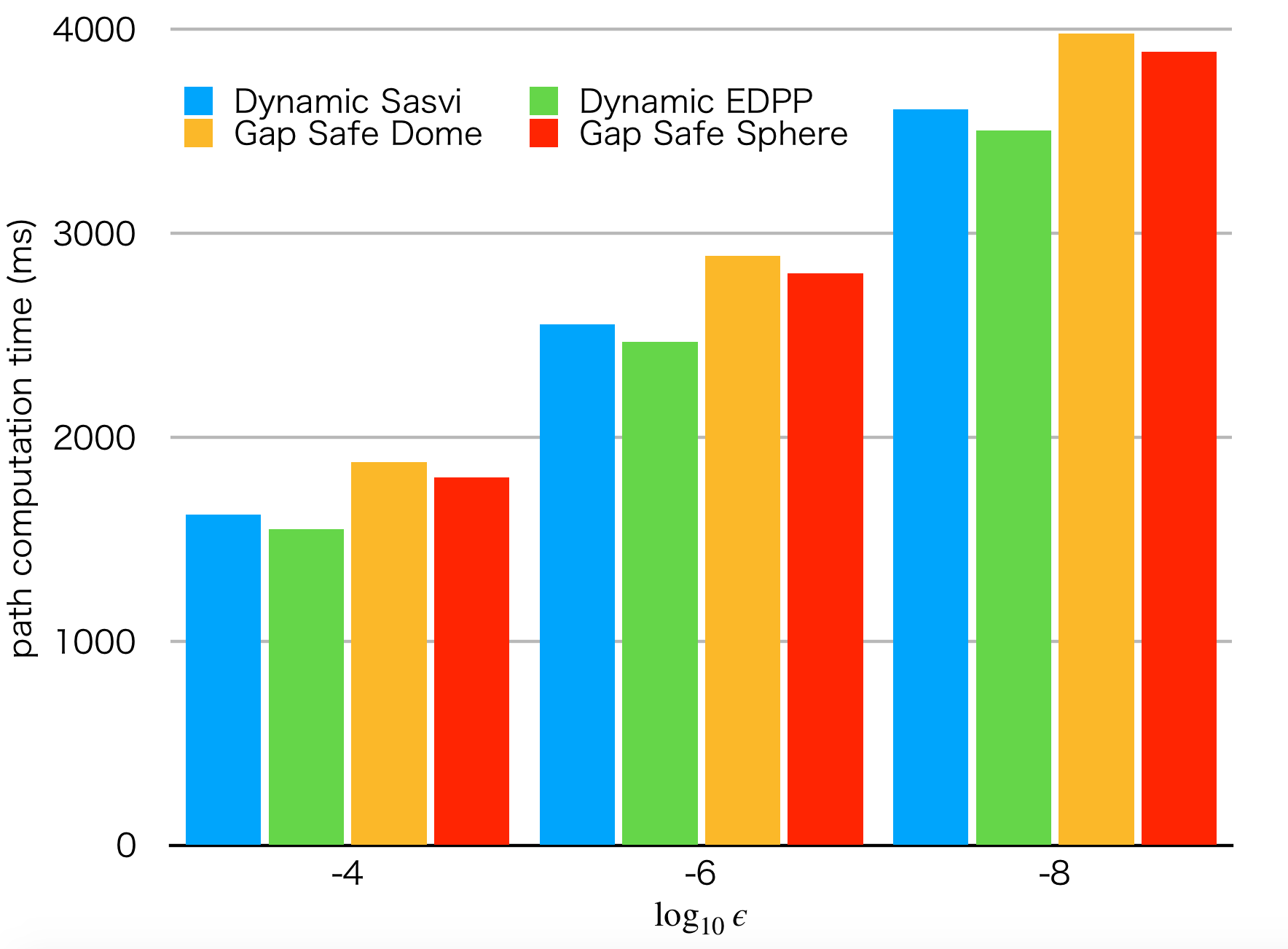}
    \caption{Computational time (20newsgroup).}
    \label{fig:time_path_20news}
  \end{subfigure}
\caption{(a): Feature remaining rate of each iteration for Lasso on Leukemia (density $n=72, d=7128$). (b) Average computational time of Lasso path on subsampled Leukemia (density of $n=50, d=7128$). (c): Average computational time of Lasso path on subsampled 20newsgroup (sparsity of $n=800, d=18571$)}
  \vspace{-.2in}
 \end{figure*}
 
 \begin{table*}[t]
\begin{center}
\caption{Logarithm of acceleration ratio for Leukemia and 20newsgroup. The smaller values indicate a greater speed up.}
\label{tb:logtimeratio}
\begin{tabular}{l|l|c|c|c|c|}
\hline
Dataset &-log epsilon & Dynamic Sasvi & Dynamic EDPP & Gap Safe Dome & Gap Safe Sphere \\ \hline
\multirow{3}{*}{Leukemia}& 4 & $-0.468\pm0.066$ & $-0.487\pm0.066$ & $-0.349\pm0.066$ & $-0.358\pm0.060$ \\ \cline{2-6}
& 6 & $-0.828\pm0.072$ & $-0.838\pm0.067$ & $-0.719\pm0.074$ & $-0.725\pm0.072$ \\ \cline{2-6}
& 8 & $-0.987\pm0.057$ & $-0.997\pm0.056$ & $-0.902\pm0.066$ & $-0.907\pm0.066$ \\
\hline \hline
\multirow{3}{*}{20newsgroup}& 4 & $-0.338\pm0.020$ & $-0.358\pm0.025$ & $-0.274\pm0.023$ & $-0.293\pm0.024$ \\ \cline{2-6}
& 6 & $-0.517\pm0.022$ & $-0.532\pm0.022$ & $-0.463\pm0.023$ & $-0.477\pm0.025$ \\ \cline{2-6}
& 8 & $-0.604\pm0.023$ & $-0.617\pm0.025$ & $-0.562\pm0.026$ & $-0.572\pm0.024$ \\
\hline
\end{tabular}
\end{center}
\end{table*}

\section{Experiments}
In this section, we show the efficacy of the proposed methods using real-world data.

\subsection{Setup}
We compared the proposed methods with Gap Safe Sphere and Gap Safe Dome \citep{fercoq2015mind,ndiaye2017gap}, which are state-of-the-art dynamic safe screening methods. All methods were run on a Macbook Air with a 1.1 GHz quad-core Intel Core i5 CPU with 16 GB of RAM. We implemented all methods in C++ using the Accelerate framework, which is the native framework for basic calculations.

\subsection{Number of screened variables}
First, we compared the number of screened variables among the four dynamic safe screening methods. We solved the Lasso problem using the Leukemia dataset (dense data with 72 samples and 7128 features) and $\lambda=\frac{1}{100}{\|\boldX^\top\boldy\|}_\infty$. We used cyclic coordinate descent as the iterative algorithm and screen variables for 10 iterations each.
Figure \ref{fig:feature_remaininf_rate_leukemia} shows the ratio of the uneliminated features at each iteration. As guaranteed theoretically, we can see that Dynamic Sasvi eliminates more variables in earlier steps than Gap Safe Dome and Gap Safe Sphere. The figure also shows that Dynamic EDPP, relaxed version of Dynamic Sasvi, eliminated almost the same number of features as Dynamic Sasvi.

\subsection{Gains in the computation of Lasso paths}
Next, we compared the computation time of the path of the Lasso solutions for various values of $\lambda$. Because $\lambda$ may be defined by a cross validation in practice, computing the path of the solutions is an important task. We used $\lambda_j={100}^{-\frac{j}{99}}{\|\boldX^\top\boldy\|}_\infty$ ($j=0,\dots,99$). The iterative solver stops when the duality gap is smaller than $\epsilon(P(\boldzero)-D(\boldzero))$. Note that $P(\boldzero)-D(\boldzero)$ makes the stopping criterion independent of the data scale.
We used the Leukemia and tf-idf vectorized 20newsgroup datasets (baseball versus hockey) (sparse data with 1197 samples and 18571 features). We subsampled the data 50 times and ran all methods for the same 50 subsamples. The subsampled data size is 50 for leukemia and 800 for 20newsgroup. Figures \ref{fig:time_path_leukemia} and \ref{fig:time_path_20news} show the average computation time of the Lasso path for the Leukemia dataset and 20news datasets, respectively. For all settings, dynamic Sasvi and dynamic EDPP outperform Gap Safe Dome and Gap Safe Sphere.

Table \ref{tb:logtimeratio} shows the the average and standard deviations of the logarithm of the acceleration ratio to the computational time for the same subsample without screening. Proposed methods are significantly faster than Gap Safe methods. In addition, Dynamic EDPP is a little faster than Dynamic Sasvi because the computational cost of Dynamic EDPP screening is smaller than the one of Dynamic Sasvi.

\section{Conclusion}
In this paper, we proposed a framework for safe screening based on Fenchel-Rockafellar duality and derived Dynamic Sasvi and Dynamic EDPP, which are specific safe screening methods for Lasso-like problems. Dynamic Sasvi and Dynamic EDPP can be regarded as dynamic feature elimination variants of Sasvi and EDPP, respectively. We proved that Dynamic Sasvi always eliminates more features than Gap Safe Sphere and Gap Safe Dome. Dynamic EDPP is based on the sphere relaxation of the Dynamic Sasvi region and eliminates almost the same number of features as Dynamic Sasvi. We also showed experimentally that the computational costs of the proposed methods are smaller than those of Gap Safe Sphere and Gap SafeDome.

\bibliography{example_paper}
\bibliographystyle{icml2021}

\newpage
\appendix
\onecolumn

\section{Proof of Theorems}

\subsection{Proof of Theorem \ref{theo:fenchel_rockafellar}}
\proof
According to (\citep{bauschke2011convex} Theorem 15.23), since we have assumed that
\[
\exists \boldbeta \in \mathrm{relint}(\mathrm{dom}(g))\ \mathrm{s.t.}\ \boldX\boldbeta \in \mathrm{relint}(\mathrm{dom}(f)),
\]
i.e., $\mathrm{relint}(\mathrm{dom}(f)) \cap \boldX\mathrm{relint}(\mathrm{dom}(g))$ is not empty, we have
\[
\inf_{\boldbeta\in\mathbb{R}^d} f(\boldX\boldbeta) + g(\boldbeta) = \max_{\boldtheta\in\mathbb{R}^n} - f^\star(-\boldtheta) - g^\star(\boldX^\top\boldtheta).
\]
In addition, we have assumed the existence of the optimal point. Hence, we have
\[
\min_{\boldbeta\in\mathbb{R}^d} f(\boldX\boldbeta) + g(\boldbeta) = \max_{\boldtheta\in\mathbb{R}^n} - f^\star(-\boldtheta) - g^\star(\boldX^\top\boldtheta).
\]
\proofend

\subsection{Proof of Theorem \ref{theo:region_dsasvi}}
\proof
According to Theorem \ref{theo:general_safe_region}, we can easily obtain
\[
\hat{\boldtheta} \in \calR^{\mathrm{DS}}(\tilde{\boldbeta},\tilde{\boldtheta}) := \{\boldtheta \mid l(\boldtheta;\tilde{\boldtheta}) \le u^{\mathrm{DS}}(\boldtheta;\tilde{\boldbeta})\}.
\]
By the definition of $u^{\mathrm{DS}}$, we have the following: 
\[
l(\boldtheta;\tilde{\boldtheta}) \le u^{\mathrm{DS}}(\boldtheta;\tilde{\boldbeta})
\iff l(\boldtheta;\tilde{\boldtheta}) \le -f^\star(-\boldtheta) \land 0 \le g(\tilde{\boldbeta}) - \boldtheta^\top\boldX\tilde{\boldbeta}.
\]
In addition, by Eqs.~\eqref{eq:dual_f_sqloss}, $g^\star(\boldX^\top\tilde{\boldtheta})=0$ and $L=1$, we have
\begin{align*}
l(\boldtheta;\tilde{\boldtheta}) \le -f^\star(-\boldtheta)
\iff & \frac{1}{2}{\|\boldtheta-\tilde{\boldtheta}\|}_2^2 - \frac{1}{2}{\|\tilde{\boldtheta}\|}_2^2 + \boldy^\top\tilde{\boldtheta} \le - \frac{1}{2}{\|\boldtheta\|}_2^2 + \boldy^\top\boldtheta \\
\iff & {\|\boldtheta\|}_2^2 - \boldtheta^\top(\tilde{\boldtheta}+\boldy) \le -\boldy^\top\tilde{\boldtheta} \\
\iff & {\|\boldtheta - \frac{1}{2}(\tilde{\boldtheta}+\boldy)\|}_2^2 \le \frac{1}{4}{\|(\tilde{\boldtheta}-\boldy)\|}_2^2.
\end{align*}
Hence, we have
\[
\hat{\boldtheta} \in \calR^{\mathrm{DS}}(\tilde{\boldbeta},\tilde{\boldtheta}) := \{\boldtheta \mid l(\boldtheta;\tilde{\boldtheta}) \le u^{\mathrm{DS}}(\boldtheta;\tilde{\boldbeta})\}
= \{\boldtheta \mid {\|\boldtheta - \frac{1}{2}(\tilde{\boldtheta}+\boldy)\|}_2^2 \le \frac{1}{4}{\|(\tilde{\boldtheta}-\boldy)\|}_2^2 \land 0 \le g(\tilde{\boldbeta}) - \boldtheta^\top\boldX\tilde{\boldbeta}\}.
\]
\proofend

\subsection{Proof of Theorem \ref{theo:equality_sasvi}}
\proof
First, we prove $g({\hat\boldbeta}^{(\lambda_0)}) = {\hat\boldtheta}^{(\lambda_0)\top}\boldX{\hat\boldbeta}^{(\lambda_0)}$.
Because $\boldX^\top{\hat\boldtheta}^{(\lambda_0)} \in \partial g({\hat\boldbeta}^{(\lambda_0)})$ (Proposition \ref{prop:optimal_condition}), for $\forall \boldbeta$, the inequality
\[
g({\hat\boldbeta}^{(\lambda_0)}) + {\hat\boldtheta}^{(\lambda_0)\top}\boldX(\boldbeta-{\hat\boldbeta}^{(\lambda_0)}) \le g(\boldbeta)
\]
holds. We can set $\boldbeta=\boldzero$ and $\boldbeta=2{\hat\boldbeta}^{(\lambda_0)}$ and obtain
\begin{align*}
g({\hat\boldbeta}^{(\lambda_0)}) - {\hat\boldtheta}^{(\lambda_0)\top}\boldX{\hat\boldbeta}^{(\lambda_0)} \le & 0 \\
g({\hat\boldbeta}^{(\lambda_0)}) + {\hat\boldtheta}^{(\lambda_0)\top}\boldX{\hat\boldbeta}^{(\lambda_0)} \le & 2g({\hat\boldbeta}^{(\lambda_0)}).
\end{align*}
Hence, we have $g({\hat\boldbeta}^{(\lambda_0)}) = {\hat\boldtheta}^{(\lambda_0)\top}\boldX{\hat\boldbeta}^{(\lambda_0)}$.

In addition, ${\hat\boldtheta}^{(\lambda_0)} = \frac{1}{\lambda_0}\boldy - \boldX{\hat\boldbeta}^{(\lambda_0)}$ holds (Proposition \ref{prop:optimal_condition}).

We then have
\begin{align*}
\calR^{\mathrm{DS}}({\hat\boldbeta}^{(\lambda_0)},{\hat\boldtheta}^{(\lambda_0)}) = & \{\boldtheta \mid {\|\boldtheta - \frac{1}{2}({\hat\boldtheta}^{(\lambda_0)}+\boldy)\|}_2^2 \le \frac{1}{4}{\|({\hat\boldtheta}^{(\lambda_0)}-\boldy)\|}_2^2 \land 0 \le g({\hat\boldbeta}^{(\lambda_0)}) - \boldtheta^\top\boldX{\hat\boldbeta}^{(\lambda_0)}\} \\
= & \{\boldtheta \mid {\|\boldtheta\|}_2^2 - \boldtheta^\top({\hat\boldtheta}^{(\lambda_0)}+\boldy) + \frac{1}{4}{\|({\hat\boldtheta}^{(\lambda_0)}+\boldy)\|}_2^2 \le \frac{1}{4}{\|({\hat\boldtheta}^{(\lambda_0)}-\boldy)\|}_2^2 \land 0 \le {({\hat\boldtheta}^{(\lambda_0)}-\boldtheta)}^\top\boldX{\hat\boldbeta}^{(\lambda_0)}\} \\
= & \{\boldtheta \mid {\|\boldtheta\|}_2^2 - \boldtheta^\top({\hat\boldtheta}^{(\lambda_0)}+\boldy) + \boldy^\top{\hat\boldtheta}^{(\lambda_0)} \le 0 \land 0 \le {({\hat\boldtheta}^{(\lambda_0)}-\boldtheta)}^\top(\frac{1}{\lambda_0}\boldy-{\hat\boldtheta}^{(\lambda_0)})\} \\
= & \{\boldtheta \mid {(\boldy-\boldtheta)}^\top({\hat\boldtheta}^{(\lambda_0)}-\boldtheta) \le 0 \land {(\frac{1}{\lambda_0}\boldy-{\hat\boldtheta}^{(\lambda_0)})}^\top(\boldtheta-{\hat\boldtheta}^{(\lambda_0)}) \le 0\} \\
= & \calR^{\mathrm{Sasvi}}(1,\lambda_0) = \calR^{\mathrm{Sasvi}}(\lambda_0)
\end{align*}
\proofend

\subsection{Proof of Theorem \ref{theo:upperbound_gm}}
\proof
By Eq.~\eqref{eq:dual_g_normlike}, we have $D(\boldtheta) \in \{ -\infty, -f^\star(-\boldtheta) \}$. Clearly, the inequality $D(\boldtheta) \le -f^\star(-\boldtheta)$ always holds. If $-f^\star(-\boldtheta) > P(\tilde{\boldbeta})$, $D(\boldtheta)$ must be $-\infty$ because $D(\boldtheta) \le P(\tilde{\boldbeta})$. Hence, we have $D(\boldtheta) \le u^{\mathrm{GM}}(\boldtheta; \tilde{\boldbeta})$.
According to Theorem \ref{theo:general_safe_region}, we have
\begin{align*}
\hat{\boldtheta} \in & \{ \boldtheta \mid l(\boldtheta; \tilde{\boldtheta}) \le u^{\mathrm{GM}}(\boldtheta; \tilde{\boldbeta}) \} \\
= & \{ \boldtheta \mid -f^\star(-\boldtheta) \le P(\tilde{\boldbeta}) \land \frac{1}{2}{\|\boldtheta-\tilde{\boldtheta}\|}_2^2 - f^\star(-\tilde{\boldtheta}) - g^\star(\boldX^\top\tilde{\boldtheta}) \le -f^\star(-\boldtheta) \} \\
= & \{ \boldtheta \mid -f^\star(-\boldtheta) \le P(\tilde{\boldbeta}) \land \frac{1}{2}{\|\boldtheta-\tilde{\boldtheta}\|}_2^2 - \frac{1}{2}{\|\tilde{\boldtheta}\|}_2^2 + \boldy^\top\tilde{\boldtheta} \le  - \frac{1}{2}{\|\boldtheta\|}_2^2 + \boldy^\top\boldtheta \} \\
= & \{ \boldtheta \mid -f^\star(-\boldtheta) \le P(\tilde{\boldbeta}) \land {\|\boldtheta\|}_2^2 - \boldtheta^\top\tilde{\boldtheta} + \boldy^\top\tilde{\boldtheta} \le \boldy^\top\boldtheta \} \\
= & \{ \boldtheta \mid -f^\star(-\boldtheta) \le P(\tilde{\boldbeta}) \land {(\tilde{\boldtheta}-\boldtheta)}^\top(\boldy-\boldtheta) \le 0 \}
\end{align*}
\proofend

\subsection{Proof of Theorem \ref{theo:region_dynamicedpp}}

\noindent{\bf Proof of $\calR^{\mathrm{DS}}(\tilde{\boldbeta},\tilde{\boldtheta}) \subset \calR^{\mathrm{DE}}(\tilde{\boldbeta},\tilde{\boldtheta})$:}

\proof
Since $\alpha\ge0$, we have
\begin{align*}
\calR^{\mathrm{DS}}(\tilde{\boldbeta},\tilde{\boldtheta}) = & \{ \boldtheta \mid {\|\boldtheta - \frac{1}{2}(\tilde{\boldtheta}+\boldy)\|}_2^2 \le \frac{1}{4}{\|(\tilde{\boldtheta}-\boldy)\|}_2^2 \land 0 \le g(\tilde{\boldbeta}) - \boldtheta^\top\boldX\tilde{\boldbeta} \} \\
\subset & \{ \boldtheta \mid {\|\boldtheta - \frac{1}{2}(\tilde{\boldtheta}+\boldy)\|}_2^2 \le \frac{1}{4}{\|(\tilde{\boldtheta}-\boldy)\|}_2^2 + 2\alpha(g(\tilde{\boldbeta}) - \boldtheta^\top\boldX\tilde{\boldbeta}) \} \\
= & \{ \boldtheta \mid {\|\boldtheta\|}_2^2 - \boldtheta^\top(\tilde{\boldtheta}+\boldy) + 2\alpha\boldtheta^\top\boldX\tilde{\boldbeta} \le \frac{1}{4}{\|(\tilde{\boldtheta}-\boldy)\|}_2^2 - \frac{1}{4}{\|\tilde{\boldtheta}+\boldy\|}_2^2 + 2\alpha g(\tilde{\boldbeta}) \} \\
= & \{ \boldtheta \mid {\|\boldtheta - \frac{1}{2}(\tilde{\boldtheta}+\boldy) + \alpha\boldX\tilde{\boldbeta}\|}_2^2 \le \frac{1}{4}{\|(\tilde{\boldtheta}-\boldy)\|}_2^2 - \frac{1}{4}{\|\tilde{\boldtheta}+\boldy\|}_2^2 + {\|\frac{1}{2}(\tilde{\boldtheta}+\boldy) - \alpha\boldX\tilde{\boldbeta}\|}_2^2 + 2\alpha g(\tilde{\boldbeta}) \} \\
= & \{ \boldtheta \mid {\|\boldtheta - \frac{1}{2}(\tilde{\boldtheta}+\boldy) + \alpha\boldX\tilde{\boldbeta}\|}_2^2 \le \frac{1}{4}{\|(\tilde{\boldtheta}-\boldy)\|}_2^2 + \alpha^2{\|\boldX\tilde{\boldbeta}\|}_2^2 - \alpha{(\tilde{\boldtheta}+\boldy)}^\top\boldX\tilde{\boldbeta} + 2\alpha g(\tilde{\boldbeta}) \}.
\end{align*}
And by $\alpha \in \{ 0, \frac{1}{{\|\boldX\tilde{\boldbeta}\|}_2^2}(\frac{1}{2}{(\tilde{\boldtheta}+\boldy)}^\top\boldX\tilde{\boldbeta} - g(\tilde{\boldbeta})) \}$, we have
\[
\alpha^2{\|\boldX\tilde{\boldbeta}\|}_2^2 - \alpha{(\tilde{\boldtheta}+\boldy)}^\top\boldX\tilde{\boldbeta} + 2\alpha g(\tilde{\boldbeta}) = -\alpha^2{\|\boldX\tilde{\boldbeta}\|}_2^2.
\]
Hence,
\begin{align*}
\calR^{\mathrm{DS}}(\tilde{\boldbeta},\tilde{\boldtheta}) \subset & \{ \boldtheta \mid {\|\boldtheta - \frac{1}{2}(\tilde{\boldtheta}+\boldy) + \alpha\boldX\tilde{\boldbeta}\|}_2^2 \le \frac{1}{4}{\|(\tilde{\boldtheta}-\boldy)\|}_2^2 - \alpha^2{\|\boldX\tilde{\boldbeta}\|}_2^2 \} \\
= & \{ \boldtheta \mid {\|\boldtheta - \boldtheta_c\|}_2^2 \le r^2 \} \\
= & \calR^{\mathrm{DE}}(\tilde{\boldbeta},\tilde{\boldtheta})
\end{align*}
holds.
\proofend

\noindent{\bf Proof of minimality of the radius:}

\proof
Let $\boldv\in\mathbb{R}^n$ be a vector which satisfies $\boldv^\top\boldX\tilde{\boldbeta}=0$ and $\boldv^\top\boldv=1$.
Note that such a vector exists if $n\ge2$.
Then, we have $\boldtheta_c \pm r\boldv \in \calR^{\mathrm{DS}}(\tilde{\boldbeta},\tilde{\boldtheta})$ because
\[
{(\boldtheta_c \pm r\boldv)}^\top\boldX\tilde{\boldbeta} = \frac{1}{2}{(\tilde{\boldtheta}+\boldy)}^\top\boldX\tilde{\boldbeta} - \max(0, \frac{1}{2}{(\tilde{\boldtheta}+\boldy)}^\top\boldX\tilde{\boldbeta} - g(\tilde{\boldbeta})) \le g(\tilde{\boldbeta})
\]
and
\[
{\|\boldtheta_c \pm r\boldv - \frac{1}{2}(\tilde{\boldtheta}+\boldy)\|}_2^2 = {\|-\alpha\boldX\tilde{\boldbeta} \pm r\boldv\|}_2^2 = \frac{1}{4}{\|(\tilde{\boldtheta}-\boldy)\|}_2^2
\]
hold.
Since the distance between these two points is $2r$, the radius of a sphere which includes $\calR^{\mathrm{DS}}(\tilde{\boldbeta},\tilde{\boldtheta})$ can not be smaller than $r$.
\proofend

\section{Direct Expression of $\max_{\boldtheta\in\calR^{\mathrm{DS}}(\tilde{\boldbeta},\tilde{\boldtheta})} \boldx_j^\top\boldtheta$}

Let $r = \frac{1}{2}{\|\tilde{\boldtheta}-\boldy\|}_2$ and $\boldtheta_o = \frac{1}{2}(\tilde{\boldtheta}+\boldy)$.

If ${(\boldtheta_o + \frac{r}{{\|\boldx_j\|}_2}\boldx_j)}^\top\boldX\tilde{\boldbeta} \le g(\tilde{\boldbeta})$, $\argmax_{\boldtheta\in\calR^{\mathrm{DS}}(\tilde{\boldbeta},\tilde{\boldtheta})} \boldx_j^\top\boldtheta
= \boldtheta_o + \frac{r}{{\|\boldx_j\|}_2}\boldx_j$ and $\max_{\boldtheta\in\calR^{\mathrm{DS}}(\tilde{\boldbeta},\tilde{\boldtheta})} \boldx_j^\top\boldtheta = \boldx_j^\top\boldtheta_o + r{\|\boldx_j\|}_2$.

If ${(\boldtheta_o + \frac{r}{{\|\boldx_j\|}_2}\boldx_j)}^\top\boldX\tilde{\boldbeta} > g(\tilde{\boldbeta})$, the constraint $\boldtheta^\top\boldX\tilde{\boldbeta} \le g(\tilde{\boldbeta})$ guaranteed to be active at the solution. Hence, we have
\begin{align*}
& \max_{\boldtheta\in\calR^{\mathrm{DS}}(\tilde{\boldbeta},\tilde{\boldtheta})} \boldx_j^\top\boldtheta \\
= & \max_{{\|\boldtheta - \boldtheta_o\|}_2^2 \le r^2 \land \boldtheta^\top\boldX\tilde{\boldbeta} = g(\tilde{\boldbeta})} \boldx_j^\top\boldtheta \\
= & \boldx_j^\top\boldtheta_o + \frac{\boldx_j^\top\boldX\tilde{\boldbeta}}{{\|\boldX\tilde{\boldbeta}\|}_2^2}(g(\tilde{\boldbeta}) - \boldtheta_o^\top\boldX\tilde{\boldbeta}) + \max_{{\|\boldtheta'\|}_2^2 \le r^2 \land {\boldtheta'}^\top\boldX\tilde{\boldbeta} = g(\tilde{\boldbeta}) - \boldtheta_o^\top\boldX\tilde{\boldbeta}} {\left(\boldx_j-\frac{\boldx_j^\top\boldX\tilde{\boldbeta}}{{\|\boldX\tilde{\boldbeta}\|}_2^2}\boldX\tilde{\boldbeta}\right)}^\top\boldtheta' \\
= & \boldx_j^\top\boldtheta_o + \frac{\boldx_j^\top\boldX\tilde{\boldbeta}}{{\|\boldX\tilde{\boldbeta}\|}_2^2}(g(\tilde{\boldbeta}) - \boldtheta_o^\top\boldX\tilde{\boldbeta}) + {\left\|\boldx_j-\frac{\boldx_j^\top\boldX\tilde{\boldbeta}}{{\|\boldX\tilde{\boldbeta}\|}_2^2}\boldX\tilde{\boldbeta}\right\|}_2\sqrt{r^2-\frac{1}{{\|\boldX\tilde{\boldbeta}\|}_2^2}{(g(\tilde{\boldbeta}) - \boldtheta_o^\top\boldX\tilde{\boldbeta})}^2}.
\end{align*}
Let $\delta=g(\tilde{\boldbeta}) - \boldtheta_o^\top\boldX\tilde{\boldbeta}$. we then have
\[
\max_{\boldtheta\in\calR^{\mathrm{DS}}(\tilde{\boldbeta},\tilde{\boldtheta})} \boldx_j^\top\boldtheta = \begin{cases}
\boldx_j^\top\boldtheta_o + r{\|\boldx_j\|}_2 & (\frac{r}{{\|\boldx_j\|}_2}\boldx_j^\top\boldX\tilde{\boldbeta} \le \delta) \\
\boldx_j^\top\boldtheta_o + \frac{\boldx_j^\top\boldX\tilde{\boldbeta}}{{\|\boldX\tilde{\boldbeta}\|}_2^2}\delta + {\left\|\boldx_j-\frac{\boldx_j^\top\boldX\tilde{\boldbeta}}{{\|\boldX\tilde{\boldbeta}\|}_2^2}\boldX\tilde{\boldbeta}\right\|}_2\sqrt{r^2-\frac{1}{{\|\boldX\tilde{\boldbeta}\|}_2^2}\delta^2} & (\frac{r}{{\|\boldx_j\|}_2}\boldx_j^\top\boldX\tilde{\boldbeta} > \delta)
\end{cases}
\]

\section{Regions for other problems}

According to Theorem \ref{theo:general_safe_region}, we can construct simple safe region by constructing simple upperbound of $D(\boldtheta)$. Herein, we introduce some regions for non Lasso-like problems.

\noindent {\bf Elastic-Net}:
Consider the following problem:
\[
\underset{\boldbeta\in\mathbb{R}^d}{\mathrm{minimize}}~~ \frac{1}{2}{\|\boldy-\boldX\boldbeta\|}_2^2 + g(\boldbeta),
\]
where $g(\boldbeta) = {\|\boldbeta\|}_1 + \frac{\gamma}{2}{\|\boldbeta\|}_2^2$ and $\gamma>0$.
Then, for $\forall \tilde{\boldbeta}$, we have
\begin{align*}
- g^\star(\boldX^\top\boldtheta) \le &  \inf_{k\ge0} g(k\tilde{\boldbeta}) - \boldtheta^\top\boldX(k\tilde{\boldbeta}) \\
= &  \inf_{k\ge0} k({\|\tilde{\boldbeta}\|}_1-\boldtheta^\top\boldX\tilde{\boldbeta}) + \frac{\gamma k^2}{2}{\|\tilde{\boldbeta}\|}_2^2 \\
= & \begin{cases}
0 & ({\|\tilde{\boldbeta}\|}_1-\boldtheta^\top\boldX\tilde{\boldbeta} \ge 0) \\
-\frac{{({\|\tilde{\boldbeta}\|}_1-\boldtheta^\top\boldX\tilde{\boldbeta})}^2}{2\gamma{\|\tilde{\boldbeta}\|}_2^2} & ({\|\tilde{\boldbeta}\|}_1-\boldtheta^\top\boldX\tilde{\boldbeta} < 0). \\
\end{cases}
\end{align*}
Because this is stronger than the Fenchel–Young inequality in Eq.~\eqref{eq:fenchel_young}, the region derived from it and Eq.~\eqref{eq:dual_f_sqloss},
\[
\left\{ \boldtheta \mid l(\boldtheta;\tilde{\boldtheta}) \le -\frac{1}{2}{\|\boldtheta-\boldy\|}_2^2 + \frac{1}{2}{\|\boldy\|}_2^2 - \frac{{\min(0,{\|\tilde{\boldbeta}\|}_1-\boldtheta^\top\boldX\tilde{\boldbeta})}^2}{2\gamma{\|\tilde{\boldbeta}\|}_2^2} \right\},
\]
is narrower than the region of Gap Safe Sphere. Since this region is a little complex, we propose to use the sphere relaxation.

\noindent{\bf General regularized least squares:}
Except for Elastic-Net, there are many regularizers that do not satisfy Eq.~\eqref{eq:g_normlike}, e.g., squared L1 regularization.
In addition, the dual problem of SVM can be seen as the regularized least squares.
In those cases, we propose using the upper bound
\[
D(\boldtheta) \le - \frac{1}{2}{\|\boldtheta-\boldy\|}_2^2 + \frac{1}{2}{\|\boldy\|}_2^2 + g(\tilde{\boldbeta}) - \boldtheta^\top\boldX\tilde{\boldbeta}.
\]
This is based on the Fenchel–Young inequality for $g$ and Eq.~\eqref{eq:dual_f_sqloss}. Note that the region
\[
\{ \boldtheta \mid l(\boldtheta;\tilde{\boldtheta}) \le - \frac{1}{2}{\|\boldtheta-\boldy\|}_2^2 + \frac{1}{2}{\|\boldy\|}_2^2 + g(\tilde{\boldbeta}) - \boldtheta^\top\boldX\tilde{\boldbeta}\}
\]
is a sphere in the Gap Safe Sphere region.

\noindent {\bf General norm regularized problems:} Here, we extend $f$ to a more general setup, e.g., the logistic loss.
Assume that $g$ satisfies Eq.~\eqref{eq:g_normlike}.
In those cases, we propose using
\begin{align*}
D(\boldtheta) & \le f(\boldX\boldbeta) + \boldtheta^\top\boldX\boldbeta + \inf_{k\ge0}g(k\tilde{\boldbeta}) - \boldtheta^\top\boldX(k\tilde{\boldbeta}), \\
& = f(\boldX\boldbeta) + \boldtheta^\top\boldX\boldbeta + \begin{cases}
0 & (g(\tilde{\boldbeta}) - \boldtheta^\top\boldX\tilde{\boldbeta} \ge 0) \\
-\infty & (g(\tilde{\boldbeta}) - \boldtheta^\top\boldX\tilde{\boldbeta} < 0)
\end{cases}.
\end{align*}
This is based on the Fenchel–Young inequality for $f$ and Eq.~\eqref{eq:gdual_linear_const}.

\end{document}

%% file: mathdef.tex
\newtheorem{defi}{Definition}
\newtheorem{prop}[defi]{Proposition}
\newtheorem{lemm}[defi]{Lemma}
\newtheorem{theo}[defi]{Theorem}
\newtheorem{coro}[defi]{Corollary}

\newcommand{\proof}{\noindent\textbf{(Proof)} }
\newcommand{\proofend}{\hfill$\Box$\vspace{2mm}}

\newcommand{\argmin}{\mathop{\mathrm{argmin\,}}}
\newcommand{\argmax}{\mathop{\mathrm{argmax\,}}}

\newcommand{\boldzero}{{\boldsymbol{0}}}
\newcommand{\boldone}{{\boldsymbol{1}}}

\newcommand{\boldX}{{\boldsymbol{X}}}

\newcommand{\boldv}{{\boldsymbol{v}}}
\newcommand{\boldw}{{\boldsymbol{w}}}
\newcommand{\boldx}{{\boldsymbol{x}}}
\newcommand{\boldy}{{\boldsymbol{y}}}
\newcommand{\boldz}{{\boldsymbol{z}}}

\newcommand{\boldbeta}{{\boldsymbol{\beta}}}

\newcommand{\boldtheta}{{\boldsymbol{\theta}}}

\newcommand{\calA}{{\mathcal{A}}}

\newcommand{\calR}{{\mathcal{R}}}

